\pdfoutput=1
\documentclass[11pt]{article}

\usepackage[final]{acl}

\usepackage{times}
\usepackage{latexsym}
\usepackage{appendix}

\usepackage{xcolor}
\usepackage{amssymb}
\usepackage[T1]{fontenc}
\usepackage[utf8]{inputenc}
\usepackage{wrapfig}
\usepackage{microtype}

\usepackage{inconsolata}
\usepackage{amsmath}
\usepackage{array}
\usepackage{graphicx}
\usepackage{titling}
\usepackage{algorithm}
\usepackage{algpseudocode}
\usepackage{booktabs}
\usepackage{marvosym}
\usepackage{amsmath}
\usepackage{amssymb}
\title{CoMet: Metaphor-Driven Covert Communication \\ for Multi-Agent Language Games}

\author{
    Shuhang Xu\textsuperscript{$\diamondsuit$$\dagger$} \and Fangwei Zhong\textsuperscript{$\diamondsuit$$\dagger$\Letter},\\
    \textsuperscript{$\diamondsuit$} School of Artificial Intelligence, Beijing Normal University, Beijing, China\\
    \textsuperscript{$\dagger$} Engineering Research Center of Intelligent Technology and Educational Application, \\
    Ministry of Education, Beijing, China\\
    \Letter Correspondence to: \textit{fangweizhong@bnu.edu.cn}\\
}

\begin{document}

\maketitle

\begin{figure*}[h!]
    \centering
    \includegraphics[width=1\textwidth]{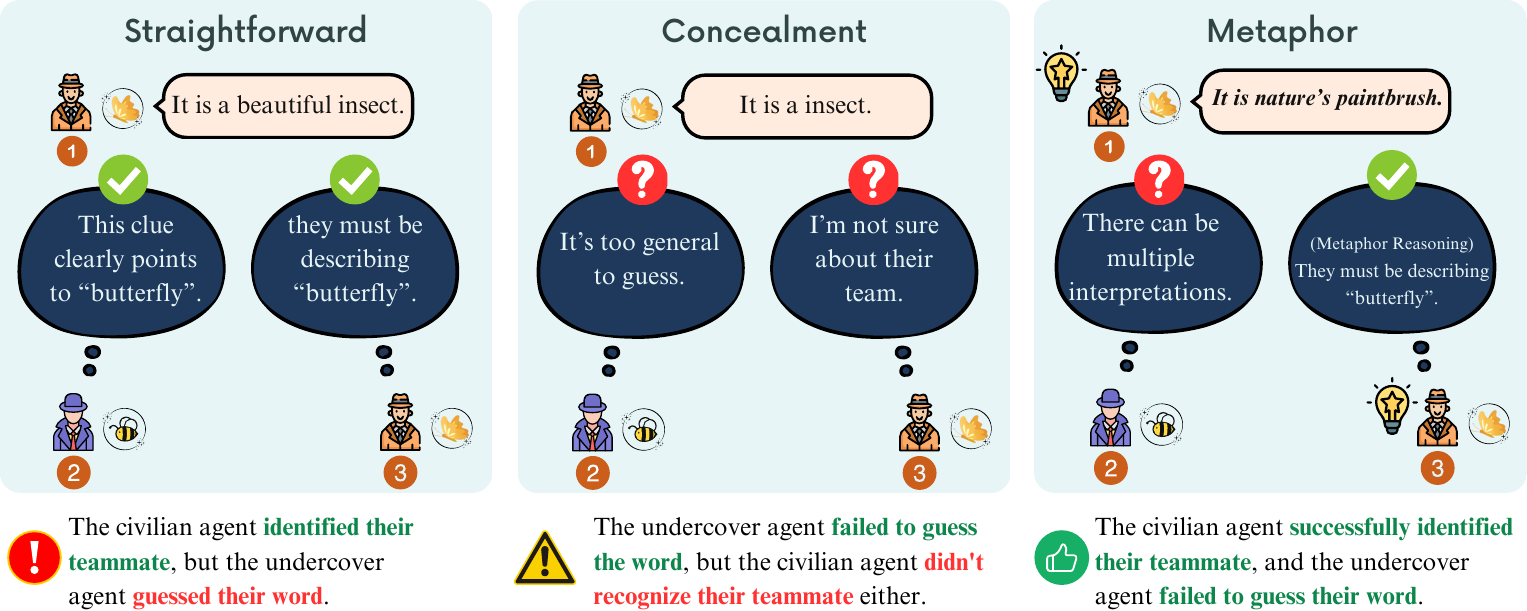} 
    \caption{
       Comparison of three communication strategies—Straightforward Description, Concealment, and Metaphorical Description—in \textit{Undercover}. In this example, a civilian describes a ``butterfly'', and the reactions of the two players are shown. In the Straightforward method, the civilian successfully identifies their teammate, but the undercover agent guesses the word. In Concealment, the civilian’s vague clue leads to confusion, with the undercover agent failing to guess the word and the civilian unable to identify their teammate. The Metaphor method allows the civilian to subtly describe the word, leading to a correct identification by the civilian agent, while the undercover agent fails to guess the word.
    % The different effects of three communication methods—Straightforward Description, Concealment, and Using Metaphors—in \textit{Undercover}. A civilian is describing a ``butterfly'' and the other two players show their reactions. Using metaphors allows the civilians to convey information secretly without being interpreted by the undercover.
    }
    \label{graph1}
\end{figure*}

\begin{abstract}
Metaphors are a crucial way for humans to express complex or subtle ideas by comparing one concept to another, often from a different domain. However, many large language models (LLMs) struggle to interpret and apply metaphors in multi-agent language games, hindering their ability to engage in covert communication and semantic evasion, which are crucial for strategic communication. To address this challenge, we introduce CoMet, a framework that enables LLM-based agents to engage in metaphor processing. CoMet combines a hypothesis-based metaphor reasoner with a metaphor generator that improves through self-reflection and knowledge integration. This enhances the agents' ability to interpret and apply metaphors, improving the strategic and nuanced quality of their interactions. We evaluate CoMet on two multi-agent language games—\textit{Undercover} and \textit{Adversarial Taboo}—which emphasize \textit{``covert communication''} and \textit{``semantic evasion''}. Experimental results demonstrate that CoMet significantly enhances the agents' ability to communicate strategically using metaphors.

%\wj{
%Metaphors are a crucial way for humans to express complex or subtle ideas by comparing one concept to another, often from a different domain. However, many large language models (LLMs) struggle to interpret and apply metaphors in multi-agent language games, hindering their ability to engage in covert communication and semantic evasion, which are crucial for strategic interaction. To address this challenge, we introduce CoMet, a framework that enables LLM-based agents to engage in metaphorical reasoning. CoMet combines a hypothesis-based metaphor reasoner with a metaphor generator that improves through self-reflection and knowledge integration. This enhances the agents' ability to interpret and apply metaphors, improving the strategic and nuanced quality of their interactions. We evaluate CoMet on two multi-agent language games—\textit{Undercover} and \textit{Adversarial Taboo}—which emphasize \textit{``covert communication''} and \textit{``semantic evasion''}. Experimental results demonstrate that CoMet significantly enhances the agents' ability to communicate strategically using metaphors.}

\end{abstract}

\section{Introduction}

In human social cognition, metaphors transcend mere rhetorical devices to constitute fundamental building blocks of communication. The power of metaphors lies in their ability to distill intricate concepts into accessible features, enriching the communicative landscape of multi-agent interactions. In dynamic interactions, metaphors can also serve as signals, hinting at underlying intentions or emotions that might otherwise remain obscured.

The understanding and use of metaphors for communication have great value and necessity in real-life scenarios. For example, metaphors can serve as a ``natural language version of asymmetric encryption'' to protect trade secrets and personal privacy. In international negotiations, metaphorical expressions like ``weather forecasting'' can signal strategic shifts without explicit commitment, functioning as a ``weak identity verification'' tool among trusted parties. On the other hand, misunderstanding metaphors in real-world applications can lead to communication breakdowns and a poorer user experience \cite{cankao1, cankao2, cankao3, cankao4, cankao6}. Most importantly, since metaphors are intrinsic to human language, enhancing AI's ability to understand and generate metaphors can improve human-AI alignment, enabling systems to understand human language expressions more comprehensively. It is essential for achieving human-level social interactions.

% For instance, during a medical consultation, a patient might use the metaphor “My heart feels like it’s being crushed under a heavy weight” to describe chest tightness. If the medical assistant system fails to interpret this metaphor and instead suggests mental stress rather than a potential heart-related issue, the outcome could be detrimental. 
% Metaphors are vital for expressing complex intentions in multi-agent interactions. They distill intricate concepts into accessible ideas, enhancing communication. In dynamic settings, metaphors can signal underlying intentions or emotions that might otherwise remain hidden. For example, players refer to weapons as "gardening tools" to evade lexical monitoring in Undercover scenarios, and in adversarial negotiations, "weather forecasting" metaphors indicate strategic shifts without explicit commitment. Misunderstanding metaphors in real-world applications can lead to communication breakdowns and a poorer user experience. Therefore, improving AI agents' comprehension of metaphors is essential for facilitating human-like social interactions.

Recent studies have increasingly utilized large language models (LLMs) as the foundation of AI agents to communicate and interact with humans or other agents, yielding impressive results \cite{guo2024survey, xu-etal-2024-magic,li-2025-review, amadeus-etal-2024-bridging}. In addition, there has been notable progress in research on metaphor understanding and generation using LLMs \cite{kim2023metaphorian, lin2024dual, aono-etal-2024-verifying}. 

% the communication of current LLM agents predominantly remains limited to straightforward expression, with little capability for understanding and utilizing metaphors. 
However, Current LLM agents exhibit catastrophic failures in  contexts with metaphors due to literal interpretation bias.
For example, we evaluated the performance of LLM agents using two strategic language games: \textit{Undercover} \cite{xu-etal-2024-magic} and \textit{Adversarial Taboo} \cite{cheng2024selfplaying}. These games test agents' abilities to use complex communication strategies, particularly metaphors. In \textit{Undercover}, agents employ metaphors for concealment and deception, a concept we term ``concept camouflage.'' In \textit{Adversarial Taboo}, the agents need to bypass forbidden words through reasoning and misdirection, addressing the ``semantic avoidance'' challenge. Our evaluation reveals that LLM agents, lacking metaphorical reasoning capabilities, struggle to implement these strategies effectively. 
% As a result, undercover agents fail to conceal their identities in \textit{Undercover}, while speakers in \textit{Adversarial Taboo} lack the subtlety needed to avoid revealing critical information.

To overcome these limitations, we introduce CoMet, a framework designed to enhance LLMs’ ability to reason with and generate metaphors. CoMet integrates two key components: a metaphor reasoning module based on hypothesis testing, and a metaphor generation module that leverages knowledge injection and experience accumulation for continuous self-improvement. The metaphor reasoning module enables the agent to understand and expand metaphors for covert communication, and the metaphor generator produces strategic, context-sensitive speech for effective communication in multi-agent games. We tested CoMet on two multi-agent language games: \textit{Undercover} and \textit{Adversarial Taboo}. \textit{Undercover} divides multiple players into two teams, with most players receiving the same word and a few players (undercover agents) receiving a different word. Players take turns describing words and voting to find the undercover agents, while the undercover agents try to hide their identities as much as possible. Adversarial taboos consist of attackers and defenders. Attackers need to guide defenders to say a secret word, while defenders need to guess the word. Specific game rules can be found in the appendix \ref{app:rule}.

Figure \ref{graph1} shows an example from \textit{Undercover}, where civilians use metaphors to encode communication and conceal private information that benefits the undercover agents. We conduct a thorough evaluation of the agents’ performance on both \textit{Undercover} and \textit{Adversarial Taboo}. The quantitative and qualitative results demonstrate that the use of metaphors enables LLM agents to effectively apply complex communication strategies, such as concealment, deception, and misdirection, in multi-agent language games.

Our key contributions are as follows: 
1) \textbf{Exploration of a new research direction}: We introduce the concept of using metaphors in communication-based games, aiming to expand the strategic options available to multi-agent systems and explore how metaphorical reasoning can enhance agent interactions.
2) \textbf{Effective framework}: We present CoMet, a framework designed to facilitate metaphorical reasoning and generation in agents. This framework encourages agents to adopt a range of communication strategies, including metaphor-based concealment, deception, and misdirection, to improve their performance in multi-agent language games.
3) \textbf{Experiments and resources}: We conduct a set of experiments to evaluate the performance of various LLMs on two benchmark games, \textit{Undercover} and \textit{Adversarial Taboo}, offering insights into the agents’ ability to employ metaphor-driven communication strategies. Ablation studies are included to examine the impact of each component within the framework. Additionally, we provide the code for \textit{Undercover} and a collected word dataset to facilitate further exploration and development.

\section{Related Works}

\paragraph{Metaphors in Natural Language Processing.}
The importance of metaphors in natural language processing (NLP) is widely recognized~\cite{shutova2010models,veale2022metaphor}, with extensive research focusing on metaphor detection, reasoning, generation, and dataset creation~\cite{li2023metaphor, mao-etal-2024-metapro, tong2024metaphor, reimann-scheffler-2024-metaphors,  lin2024dual, jia-li-2024-metaphor}. With the rapid advancement of large language models (LLMs), researchers have shown that LLMs can process metaphors \cite{kim2023metaphorian, tong2024metaphor,  tian2024theory, Liu2022TestingTA}. However, existing research mainly focuses on addressing static text data, while the use of metaphors in dynamic, interactive multi-agent scenarios, such as multi-agent language games, has received limited attention. This study explores the integration of metaphor understanding, reasoning, and generation into multi-agent language interactions, aiming to uncover more nuanced communication patterns during complex interactions.
% In the field of natural language processing, the importance of metaphor has reached a broad consensus, with researchers conducting in-depth explorations around metaphor detection, reasoning, generation, and datasets~\cite{mao-etal-2024-metapro, tong2024metaphor, reimann-scheffler-2024-metaphors, lin2024dual, jia-li-2024-metaphor}. The rapid development of large language models (LLMs) has also demonstrated the feasibility of utilizing LLMs to handle metaphors \cite{tong2024metaphor, kim2023metaphorian, tian2024theory}. However, existing research on metaphors has primarily focused on the analysis of static text data, while the application of metaphors in dynamic interactive scenarios involving multi-agent systems remains largely unexplored. This study first attempted to integrate metaphor mechanisms into multi-agent language interaction systems, exploring richer communication patterns among agents in dynamic collaboration.

\paragraph{Multi-Agent Language Games.}
With the advancement of LLMs, researchers have utilized language games as interactive environments to examine multi-agent interactions. These games are generally categorized into three types: adversarial games, cooperative games, and mixed games. The adversarial games, such as \textit{Diplomacy} \cite{mukobi2023diplomacy, guan2025richelieu} and \textit{Adversarial Taboo} \cite{AT}, focus on maximum agents’ self-interest through adversarial strategies. The cooperative games, such as \textit{Referential Game}~\cite{yuan2020emergence}, require agents to collaborate toward shared objectives. 
The mixed games not only cooperation among teammates but also compete against some adversaries, such as
\textit{Werewolf} \cite{xu2023werewolf}, \textit{Avalon} \cite{light2023avalonbench}, and \textit{Chameleon} \cite{xu-etal-2024-magic}.
 These language games necessitate decision-making under incomplete information, with clear victory conditions and specific action goals. \textit{Undercover} \cite{xu-etal-2024-magic} also highlights cooperation and competition but adds complexity by keeping the agent's role unknown, challenging the reasoning process further. 
  To explore covert communication, we focus on the game settings with adversaries, specifically adversarial and mixed games.
 Thus, we select \textit{Adversarial Taboo} and \textit{Undercover}, representing the adversarial and mixed games, to investigate how agents utilizing metaphorical reasoning perform across different task settings.

\paragraph{Multi-Agent Communication With LLMs.}
To enhance the capabilities of LLM-based agents in multi-agent language games, various approaches have been proposed, including reasoning-guided prompt engineering \cite{Wei2022Chain, zhao2023zeroshot, NEURIPS2023_tot}, reflection-based self-improvements \cite{light2024strategist, xu2023werewolf, cheng2024selfplaying}, and memory-augmented architectures \cite{shinn2023reflexion, chen2023walking, guan2025richelieu}, among others. Current multi-agent language games often involve both cooperation and confrontation, where agents’ speech is broadcast to both teammates and opponents, thereby constraining their communication and decision-making. Covert communication with teammates, while safeguarding private information, could gain a strategic advantage by misleading adversaries. However, the use of metaphors for covert communication in multi-agent settings has been largely unexplored.

\begin{figure*}[t]
    \centering
    \includegraphics[width=1\textwidth]{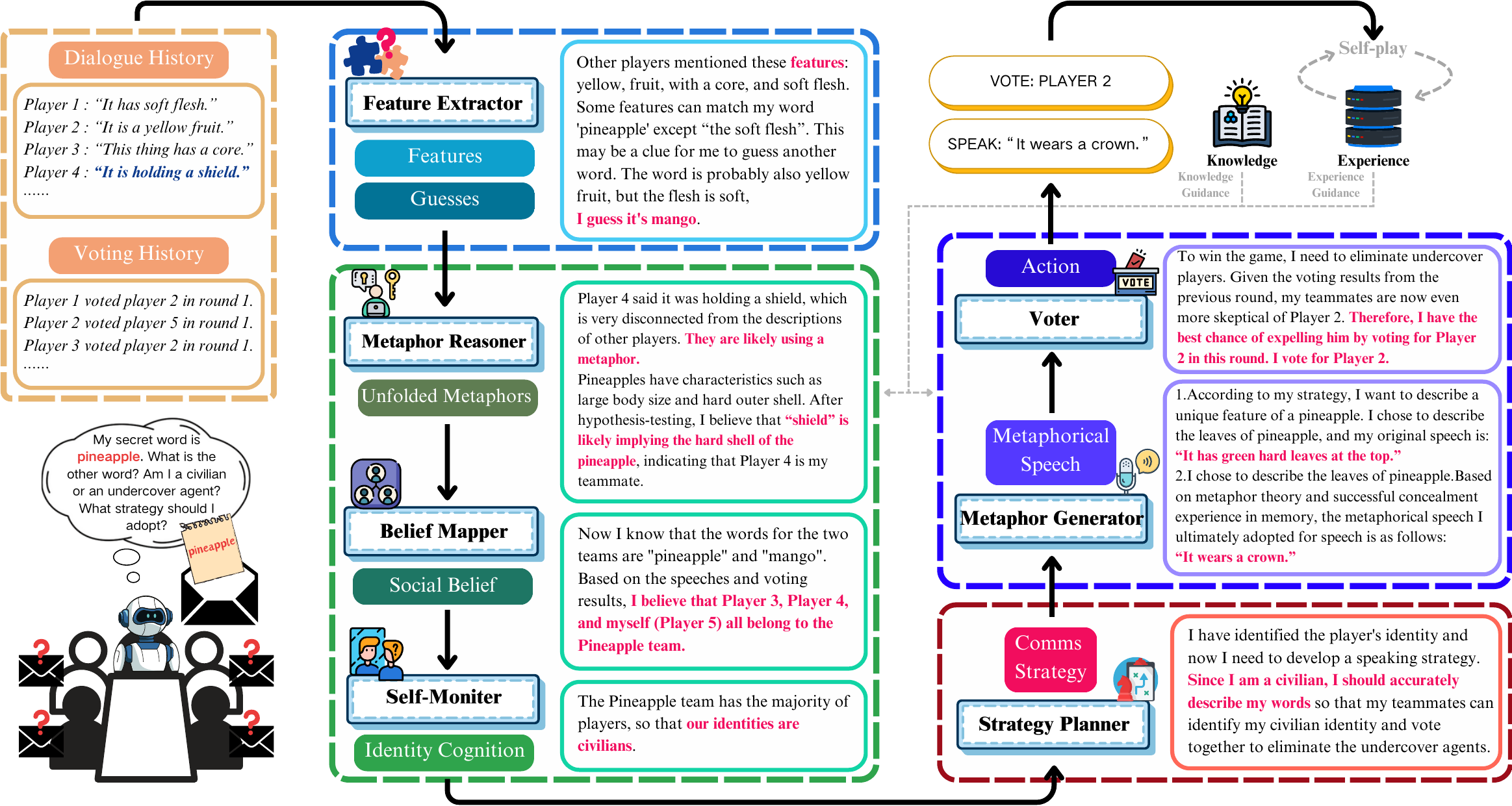} 
    \caption{
    Overview of the CoMet framework, illustrated within the ``concept camouflage” task in \textit{Undercover}. 
    % The framework consists of four key components: 1) the Feature Extractor, which processes the input data; 2) the Metaphor and Mind Reasoner, including the Hypothesis-based Metaphor Reasoner, Belief Mapper, and Self-Monitor, which enable the agent to reason and track beliefs; 3) the Strategy Planner, which designs the agent’s communication strategy; and 4) the Actor Modules, comprising the Metaphor Generator and Voter, responsible for generating metaphorical speech and voting decisions. This structure equips agents to use metaphors effectively in multi-agent interactions, enhancing strategic communication.
The agent starts by extracting features from the game state, including player behavior and available clues. The Metaphor Reasoner identifies and expands metaphors to aid in interpretation. As the game progresses, the agent uses the Belief Mapper to build beliefs about other players’ roles and tracks its own identity with the Self-Monitor. With this understanding, the Strategy Planner formulates a communication and action strategy. The agent then generates metaphorical speech through the Metaphor Generator to communicate covertly. Finally, it votes according to its assessment, while new dialogue and voting histories are recorded to inform future decisions. 
    % Our framework equips agents to use metaphors effectively in multi-agent interactions, enhancing strategic communication.
    % The CoMet framework with \textit{Undercover} as an example. It includes: Feature Extractor; Reasoning Modules - Hypothesis-based Metaphor Reasoner, Belief Mapper, Self-Monitor; Strategy Planner; Actor Modules - Metaphor Generator and Voter.
    }
    \label{framework}
\end{figure*}

\section{Metaphor-Aware LLM Agent}
% We introduce CoMet, a framework designed to enable covert communication for agents, helping them enrich and implement complex strategies such as misdirection and concealment through the use of metaphors. 

\subsection{Overview} 
\paragraph{Game Setup.}
Taking \textit{Undercover} as an example, there are $N$ players in the game. At the beginning, each player receives a secret word from a pair of similar words \( (W_1, W_2) \). 
These words are assigned to the civilian and undercover teams, with only a few players receiving the undercover word, i.e., $P_\text{Und.} \xleftarrow{\textsf{Assign}} W_1 , P_\text{Civ.} \xleftarrow{\textsf{Assign}} W_2$. Players on the same team share the same secret word, but they are unaware of their roles and teammates, as sharing the secret word is prohibited.
Players will speak in a random order during the speaking phase, and then vote simultaneously during the voting phase. As the speaking and voting phases alternate, the game progresses until a team wins. It is now player $i$'s turn ($i \in \{1, \dots, N\}$) to think and speak.
\textit{Adversarial Taboo} can be seen as a simplified two-player game in which one word is given to one player, with each player's role being known.

\paragraph{CoMet Framework.} We introduce CoMet, a framework that enables Covert Communication by using Metaphors to implement strategies like misdirection and concealment.
Figure \ref{framework} provides an overview of CoMet (Communicating with Metaphor).
The agent begins by extracting initial features $\mathcal{F}$ from their observations $O$ of other players' behaviors and speech content, through the Feature Extractor. These features are then passed to the Metaphor Reasoner, which checks for metaphors and expands their meaning through hypothesis testing. The agent next builds its beliefs~$\mathcal{M}$ about the roles of other players using the Belief Mapper. The Self-Monitor continuously tracks the agent's own identity $I$ to ensure alignment with the correct game objectives. With this understanding, the Strategy Planner formulates a comprehensive strategy $\mathcal{S}$ that includes both communication and action. The agent then generates metaphorical speech through the Metaphor Generator to communicate covertly. Finally, the agent executes the communication and action components of its strategy through the Actor, performing the actions $\mathcal{A}$ specified by the game rules to achieve its goals.

% \begin{equation}
% P_\text{Und.} \xleftarrow{\textsf{Assign}} W_1 , P_\text{Civ.} \xleftarrow{\textsf{Assign}} W_2
% \end{equation}

% \subsection{Method Overview} 
% We introduce CoMet, a framework designed to enable covert communication for agents, helping them enrich and implement complex strategies such as misdirection and concealment through the use of metaphors. Figure \ref{framework} provides an overview of CoMet (Communicating with Metaphor). \wj{To illustrate its functionality, we use the ``concept camouflage” task in \textit{Undercover}, where agents must employ metaphors for concealment and deception, testing their ability to communicate covertly while avoiding detection.}

% % The agent begins by extracting initial features from the game environment, including player behaviors and available clues, through the Feature Extractor. This information is then passed to the Metaphor Reasoner, which checks for metaphors and expands their meaning through hypothesis testing. The agent next builds its beliefs about the roles of other players using the Belief Mapper. The Self-Monitor continuously tracks the agent's own identity and role to ensure alignment with its strategy. With this understanding, the Strategy Planner formulates a comprehensive strategy that includes both communication and action. The agent then generates metaphorical speech through the Metaphor Generator to communicate covertly. Finally, the agent executes the communication and action components of its strategy through the Actor, performing the actions specified by the game rules to achieve its goals.

In the following, we detail each step of CoMet using the ``concept camouflage'' task in \textit{Undercover}, where agents employ metaphors for covert communication. The detailed prompting template for each module is introduced in Appendix~\ref{app:prompt}.

% we provide a detailed illustration of each step in CoMet, using the “concept camouflage” task in \textit{Undercover}. In this task, agents must use metaphors for concealment and deception, showing their ability to communicate covertly while avoiding detection. See Appendix~\ref{app:prompt} for the prompting template.
% The detailed prompting template for each module is introduced in Appendix~\ref{app:prompt}.

\subsection{Feature Extractor}
In multi-agent language games, agents primarily rely on the language of other players to make decisions. Storing observations of other players' speech and actions \( O_{\alpha=1}^{N} \) and filtering out valuable information \( F_i \) from the conversation is essential, and different game rules \( R \) also affect how information is shared and interpreted.
\begin{equation}
H \leftarrow H' \cup \{ O_{\alpha} \}_{\alpha=1}^{N}
\label{eq1}
\end{equation}
\begin{equation}
\mathcal{F}_i = \textsf{Extracted-Feature} \{ H , R\} 
\label{eq2}
\end{equation}
In \textit{Undercover}, all players take turns describing their words. Therefore, player $i$ needs to analyze the descriptions made by other players and extract the characteristics of the words. They will categorize the descriptions into three types: detailed descriptions of their own word, broad descriptions of their own word, and descriptions that do not match their own word.
For example, if player $i$'s word is ``pineapple'', then ``scaly rough skin'' would be a detailed description, ``yellow fruit'' would be a general description, and ``skin with red spots'' would be a description that does not match the word.
The descriptions that do not match the word essentially describe the characteristics of another word. Players gradually collect these features and, once they have built enough confidence, they guess the other word to support their subsequent actions.

\subsection{Metaphor and Belief Reasoner}

\textbf{Hypothesis-Based Metaphor Reasoner.}
This module is used to filter other players’ descriptions, checking if they contain metaphors. Suppose the agent determines that a description does not align with the focus of the current game. In that case, it will attempt to interpret it as a metaphor and uncover its underlying meaning. To enhance the effectiveness of metaphor reasoning, we employ knowledge injection and hypothesis testing. 
To be specific, we adopt a widely accepted linguistic theory of metaphors from \cite{cankao1} as knowledge input for the agents, which can assist LLMs in better metaphor reasoning. This theory classifies metaphors into ontological metaphors, structural metaphors, and spatial metaphors. The pseudocode of the reasoning process is available in Appendix~\ref{app:hypo}.

Figure \ref{hypothesis} shows an example of the hypothesis-based metaphor reasoning process. Since our framework aims to use metaphors to achieve covert communication—in \textit{Undercover}, civilians convey to their teammates ``we share the same word” without the undercover agent discovering the content of the word—the metaphor reasoning here does not require deciphering the true meaning behind the metaphor. Instead, it only needs to make a yes-or-no judgment. This method simplifies the traditional metaphor interpretation process into a binary classification mechanism, achieving the goal while significantly reducing the semantic complexity of conventional metaphorical communication.

\begin{figure}[t]
    \centering
    \includegraphics[width=0.47\textwidth]{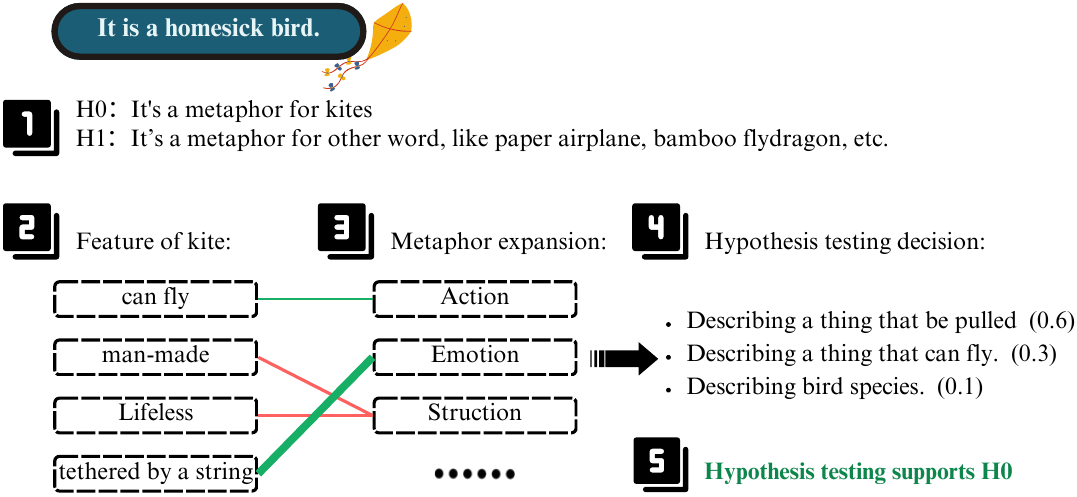} 
    \caption{The metaphor reasoning process based on hypothesis testing when players holding the word “kite” encounter the statement “homesick bird.” The process involves hypothesizing whether the metaphor refers to a kite (H0) or another object (H1), followed by analysis of features such as flight, lifelessness, and being tethered. Through metaphor expansion and hypothesis testing, the model determines that the metaphor best fits the description of a kite, supporting H0.}
    \label{hypothesis}
\end{figure}

\textbf{Belief Mapper.}
After extracting the relevant features (Eq. \ref{eq2}), the agent infers the belief of other players, denoted by \( \mathcal{M}_{-i}\) with first-order theory of mind (ToM) reasoning and the game rules. Based on the private information revealed through the received speeches, the agent will attempt to infer their identity \( I_{-i} \), role \( R_{-i} \), strategy \( S_{-i} \), and other relevant factors.
\begin{equation}
\mathcal{M}_{-i} = \{ I_{-i}, R_{-i}, S_{-i} \} = \textsf{Estimate}(\mathcal{F}_i)     
\label{eq3}
\end{equation}
In \textit{Undercover}, player $i$ will classify other players based on the categorized features: players who describe detailed characteristics of the word are considered teammates, players whose descriptions do not match the word are classified as opponents, and those who provide vague descriptions are categorized as undecided.

\textbf{Self-Monitor.}
In multi-agent language games involving identity uncertainty, it is crucial to identify one's own role based on feedback from other players. Under this module, player $i$ attempts to infer its own identity \( I_i \) by leveraging the extracted feature \( \mathcal{F}_i \) and beliefs about other players \( \mathcal{M}_-i \). 
\begin{equation}
I_i' = \textsf{Self-Awareness}(\mathcal{F}_i,\mathcal{M}_{-i})
\label{eq4}
\end{equation}
As the game progresses, the agent's understanding of its identity will be updated and refined, i.e., $I_i \leftarrow I_i'$, and the number of undecided players decreases. Once the roles of most players have been accurately inferred, player $i$ will use the game rule of ``most are civilians, few are undercover’’ to deduce their identity and clarify the objective.

\subsection{Strategy Planner}
Now it is the key module of the basic framework—we want the agent to not only analyze, reason, and make decisions, but also to employ complex communication strategies \( \mathcal{S}'_i \), such as concealment and misdirection. Since LLMs do not inherently use these methods, we need to provide the agent with guidance \( G_s \) (\( s \in S \)), helping it develop more sophisticated communication strategies. Since some strategies require multiple rounds of execution, the strategies are passed through rounds. Each time a strategy is generated, it refers to historical strategies \(\mathcal{S}_i\), and the generated strategy also provides suggestions and reminders for subsequent strategies.
\begin{equation}
\mathcal{S}'_i = \textsf{Comms-Strategy}(\mathcal{F}_i,\mathcal{M}_i, I_i , \mathcal{S}_i, G_s)
\label{eq6}
\end{equation}
\begin{equation}
\mathcal{S}_i \leftarrow \mathcal{S}'_i
\end{equation}
In the original LLM agent behavior without the CoMet framework, we found that the LLM agent, while playing \textit{Undercover}, would always directly and accurately describe its own word, leading to the exposure of all players' identities after just one round of descriptions. To address this, we require the player to adopt self-protection strategies when uncertain about their identity. At the beginning of the game, players are encouraged to describe broader and vaguer characteristics of their word to avoid revealing their identity. In each round, the player decides on their speech strategy based on the features of the word they’ve analyzed, their guesses about the other word, and their awareness of their own identity. If a player believes they are a civilian, they will balance providing details and concealing the features of their word to help teammates identify their role. However, if the player believes they are undercover and have figured out the civilian’s word, they will stop describing their own word and start describing the civilian’s word instead, attempting to deceive the opponents, blend into the civilian group, and ultimately secure a win.

\subsection{Self-improving Actor}
\textbf{Metaphor Generator.}
During the speaking phase, the agent will select the corresponding communication skills based on the established strategy and generate the content of the speech \( \mathcal{A}_i \) for this round in accordance with the game rules and the information to be conveyed.
\begin{equation}
\mathcal{A}_i=\textsf{Speak}(\mathcal{S}_i)
\label{eq7}
\end{equation}

Once the communication strategy is formulated, the agent’s speech will no longer be straightforward. Instead, it will involve deception, misdirection, or concealment, expressed through metaphors. We continue to inject relevant metaphor theories into the prompts to assist the agent in generating metaphors effectively.

Current research on LLMs and metaphors mainly focuses on detection and reasoning, while generating high-quality metaphors remains a challenge. We aim to enhance LLMs’ metaphor generation through self-play in \textit{Undercover}. By accumulating data from self-play, the agent uses game outcomes and others’ interpretations as feedback to refine its metaphor generation skills. Each metaphor creates a reference experience, including its meaning, interpretations, and suggested revisions. In future games, the agent selects relevant experiences from the reference pool to improve its prompts and generate more effective metaphors.

%\paragraph{Comms Strategy}
%Once the agent has gathered clearer information, it still needs to reason about its identity and the game state to formulate a speaking strategy. With the introduction of the Metaphor Mentor module, the agent’s range of possible behaviors increases. It can attempt to hide information or deceive others to meet its sub-goals or the final objective. When the agent chooses to use more complex speaking behaviors, the original speaking strategy is upgraded into a more advanced communication strategy, guiding subsequent actions.

\textbf{Voter.} 
% In most multi-agent language games, there are also action rounds in addition to the speaking rounds, such as the voting rounds in \textit{Undercover}. 
In \textit{Undercover} game, after the speaking round, a voting round follows, where each player votes for other players. If new observations arise between the last speech and the current vote, the agent must re-extract features, reassess the situation, and update its strategy before proceeding with voting or similar actions.

\section{Experiments}

We use two communicative language games, \textit{Adversarial Taboo} and \textit{Undercover}, as benchmarks to evaluate CoMet and other LLM-based baselines. In \textit{Undercover}, communication leans more towards conceptual descriptions, and the communication strategy focuses on concealment and encrypted conversations. In contrast, in \textit{Adversarial Taboo}, communication is more dialogue-oriented, with the communication strategy emphasizing the misleading of others. The code can be found at: \url{https://github.com/Yeswolo/CoMet}.

\subsection{Experimental Setups}

\paragraph{Adversarial Taboo} is a one-on-one competitive language game where players communicate concepts within linguistic constraints while managing adversarial interference. The \textit{attacker} has a secret word and aims to guide the \textit{defender} to say it, while the \textit{defender} attempts to avoid saying the word and collects clues to guess it. The defender wins by correctly guessing the word; if the defender fails, the attacker wins.

\paragraph{Undercover} is a structured social deduction and multi-agent language game designed to explore group dynamics, deception, and semantic reasoning. In this game, players are assigned one of two roles: \textit{Civilians}, who are given a target word (e.g., “Bicycle”), and \textit{Undercover Agents}, who are assigned a semantically related but different word (e.g., ``Motorcycle”). Players must strategically reveal hidden roles through rounds of clue-giving, communication, and voting, while avoiding detection. At the end of each round, the player with the most votes is eliminated. If there is a tie, no one is eliminated, and the game continues. Our setup includes five agents (three civilians and two undercover agents) with a maximum of 10 rounds per episode. We collected 200 word pairs across two main themes—food and animals—and each pair is tested across 10 evaluation episodes. The words we used are listed in Figures~\ref{word1} and~\ref{word2}.

\paragraph{Baseline.} The \textit{Naive} baseline is applying the LLMs to directly answer the detailed prompts. The stronger baseline is using Chain-of-Thought (CoT) \cite{Wei2022Chain} to build an agent for the two games. In the \textit{Adversarial Taboo} game, we evaluate the performance of different LLMs, including GPT-o1, DeepSeek-R1, Llama3.3-70B, Claude3.5 Sonnet and Qwen2.5-72B, using both CoT and CoMet. Both methods follow the same game rules and utilize the same in-game information. Due to the underperformance of CoT as an undercover agent, we also introduce an additional baseline by removing the metaphorical modules (Hypothesis-Based Metaphor Reasoner and Metaphor Generator) from CoMet, which we refer to as \textit{CoMet w/o Met.} in the experiments. Unless otherwise stated, GPT-4o is used as the primary LLM in the undercover experiments. Please refer to Appendix ~\ref{app:implementation} for more implementation details.

\paragraph{Evaluation Metrics.} To quantitatively assess the agents, we introduce the following metrics based on the game logs:
1) \textbf{\textit{Win Rate (WR)}} measures the agent's comprehensive performance by calculating the ratio of games won to the total number of games played.
2) \textbf{\textit{Feature Extraction Rate (FER)}} quantifies the agent's ability to capture critical features by evaluating the ratio of valid features extracted to the total speech entries received from other players.
3) \textbf{\textit{Others' Identity Assessment Accuracy (OIAA)}} reflects the agent's capability to distinguish allies from opponents, defined as the ratio of correct identity judgments to the total number of other players' speech entries.
4) \textbf{\textit{Self-Identity Assessment Accuracy (SIAA)}} evaluates the agent's consistency in maintaining its role, calculated as the ratio of successful self-identity confirmations to the total number of attempts to assess its identity.
5) \textbf{\textit{Privacy Protection Capability (PPC)}} assesses the agent's ability to safeguard private information against adversaries, expressed as subtracting the ratio of the number of leaked pieces of information to the total number of speeches from 1.
6) \textbf{\textit{Identity Inconsistent Statement Capability (IISC)}} measures the agent's strategic complexity by quantifying the frequency of deceptive or misleading statements relative to its total speech entries.
The formal definition of these metrics is introduced in Appendix \ref{app:metrics}.

We observe that agents exhibit role preferences during the game due to LLM biases, leading to inflated metrics for civilians that do not accurately reflect their true performance. Specific examples of this issue will be discussed in \ref{SIAA}. To mitigate the role bias that may arise from using the same method across different roles, we introduce \textit{Balanced Metrics}. These are calculated by first averaging the metric values for each method across both roles, and then subtracting the variance to obtain the balanced value:
$M'_i = \text{avg}(M_i^\text{Civ}, M_i^\text{Und}) - \text{Var}(M_i^\text{Civ}, M_i^\text{Und})$, 
Where \( M_i \) ($i \in \{1, \dots, 6\}$) represents the six metrics (\textit{e.g.}, WR, FER, etc.).

\subsection{Results on Adversarial Taboo Game}

\begin{figure}[t]
	\centering
	\begin{minipage}{\linewidth}
		\centering
		\includegraphics[width=0.99\linewidth]{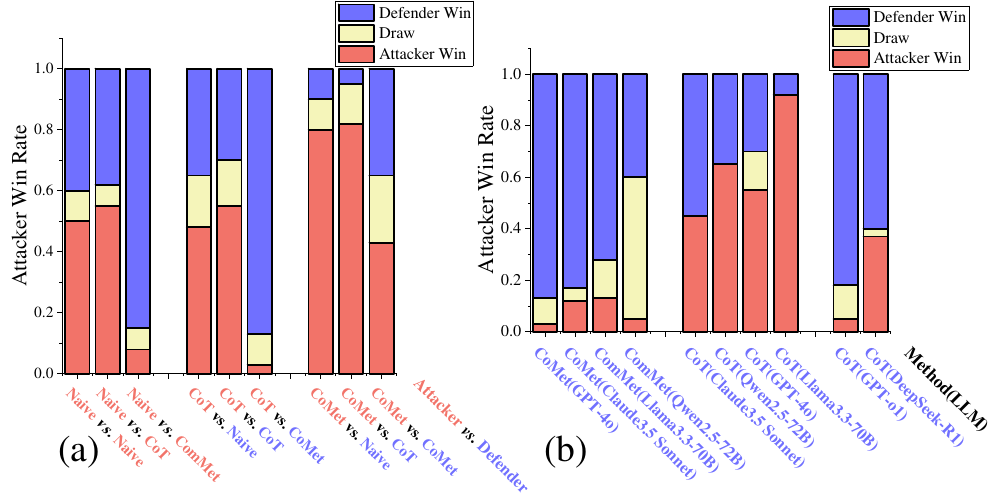}
	\end{minipage}
	%\qquad
    \caption{
    Performance comparison of different LLMs in \textit{Adversarial Taboo}. (a) Game result statistics for Naive Agent, Agent with CoT, and Agent with CoMet. (b) Performance of LLMs with various methods when facing an attacker using CoT.}
    \label{multi2}
\end{figure}

\paragraph{Playing against Baselines.} 
Figure \ref{multi2} (a) demonstrates CoMet's performance in \textit{Adversarial Taboo}, where it achieves significantly higher win rates than baseline methods both as attackers and defenders, with attackers' win rates increasing by $47\%$ and defenders' win rates increasing by $30\%$ compared to the baseline. In contrast to \textit{Undercover}, which requires cooperative covert communication through metaphors, players in \textit{Adversarial Taboo} employ metaphorical conceptual substitution to accomplish adversarial behaviors like concealment and misguidance. The results show our method's generalization capability across different games.

\paragraph{Generalization of CoMet to Different LLMs.}
Figure \ref{multi2} (b) shows the performance of different LLMs using CoT and our method CoMet. The opponent is GPT-4o using CoT. The results demonstrate that our method generalizes across different LLMs, with the use of CoMet reducing the failure rate to below $15\%$ for all tested LLMs. Specifically, GPT-4o with CoMet exhibited the best performance, achieving the highest win rate of $87\%$.

\subsection{Results on Undercover Game}

\begin{figure}[t]
    \centering
    \includegraphics[width=0.49\textwidth]{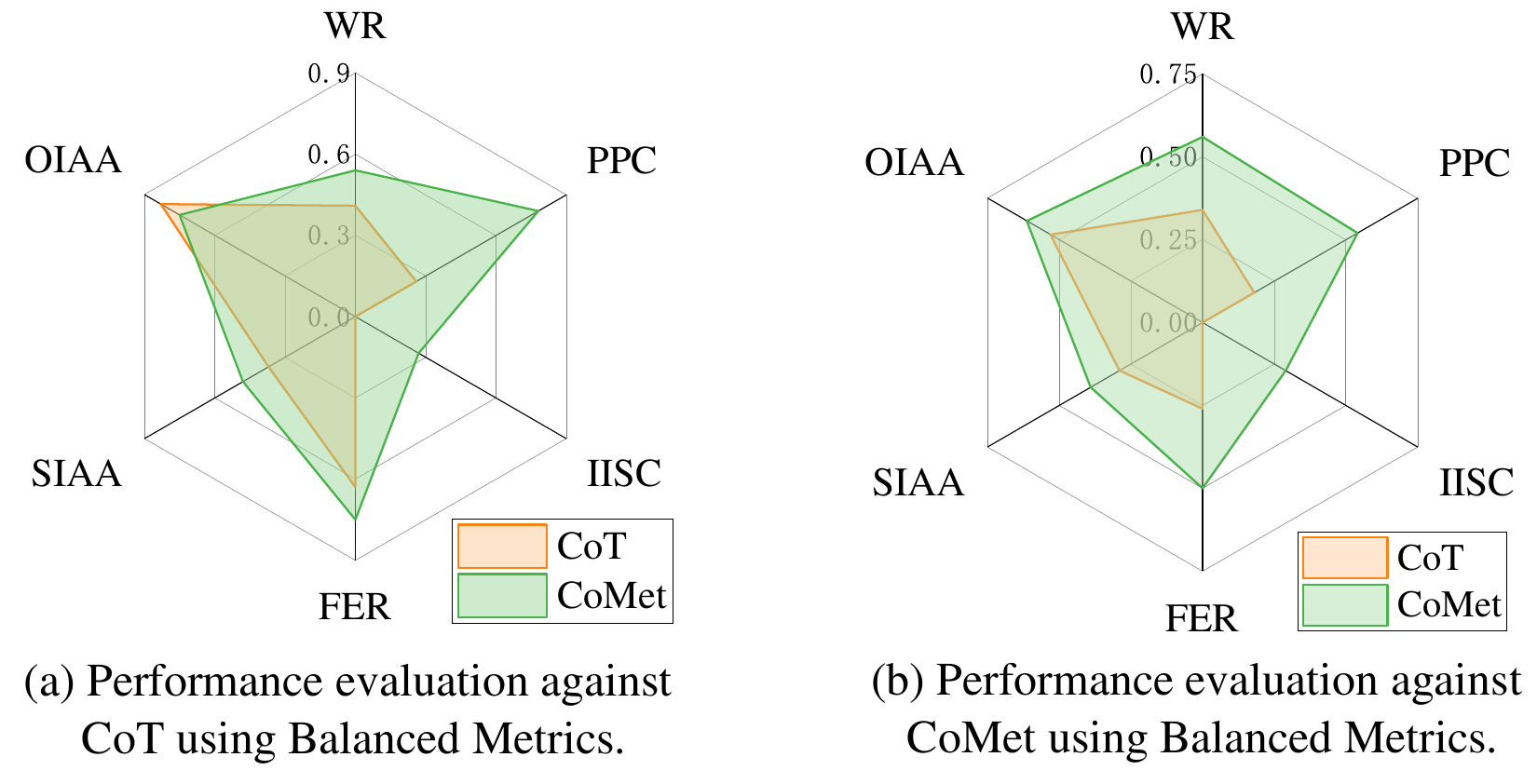} 
        \caption{Evaluation of the comprehensive performance of CoT and CoMet agents in \textit{Undercover} game using balanced metrics. 
        %Subfigures (a) and (b) show the results of each agent playing against different opponents, with (a) representing results against CoT agents and (b) against CoMet agents.
        }
        
    % Metrics include Win Rate (WR), Feature Extraction Rate (FER), Others' Identity Assessment Accuracy (OIAA), Self-Identity Assessment Accuracy (SIAA), Privacy Protection Capability (PPC) and Identity Inconsistent Statement Capability (IISC).
    \label{radar}
\end{figure}

\begin{table}[t]
\centering
\caption{Performance comparison of different methods relative to two baselines in \textit{Undercover} game, showing the results when playing two roles (undercover and civilian), where multiple players on the same team use the same method.}
\resizebox{\columnwidth}{!}{ % 确保表格适应当前栏宽度
  \begin{tabular}{lcccc | cc}
  \toprule
  \textbf{Role (Method)} & \textbf{WR}$\uparrow$ & \textbf{PPC}$\uparrow$ & \textbf{IISC}$\uparrow$ & \textbf{FER}$\uparrow$ & \textbf{SIAA} & \textbf{OIAA} \\
  \midrule
  \textcolor{gray}{against CoT} & & & & & & \\
  Und. (CoT) & 0.20 & 0.30 & 0 & 0.65 & 0.14 & \textbf{0.85} \\
  Und. (CoMet) & 0.35 & \textbf{0.82} & \textbf{0.41} & \textbf{0.77} & 0.37 & 0.74 \\
  Civ. (CoT) & 0.80 & 0.23 & 0 & 0.61 & \textbf{0.88} & 0.82 \\
  Civ. (CoMet w/o Met.) & \textbf{0.85} & 0.68 & 0.12 & 0.72 & 0.67 & \textbf{0.85} \\
  Civ. (CoMet) & \textbf{0.85} & 0.75 & 0.16 & 0.73 & 0.62 & 0.76 \\
  \midrule
  \textcolor{gray}{against CoMet w/o Met.} & & & & & & \\
  Und. (CoT) & 0.15 & 0.18 & 0 & 0.34 & 0.04 & 0.47 \\
  Und. (CoMet) & 0.45 & 0.50 & \textbf{0.37} & 0.48 & 0.31 & 0.58 \\
  Civ. (CoT) & 0.65 & 0.17 & 0 & 0.19 & \textbf{0.92} & 0.60 \\
  Civ. (CoMet w/o Met.) & 0.55 & 0.42 & 0.23 & 0.44 & 0.51 & 0.64 \\
  Civ. (CoMet) & \textbf{0.70} & \textbf{0.58} & 0.22 & \textbf{0.53} & 0.48 & \textbf{0.68} \\
  \bottomrule
  \end{tabular}
}
\label{table1}
\vspace{-0.3cm}
\end{table}

\paragraph{Playing against Baselines.} \label{SIAA}
Table \ref{table1} compares different methods based on agents’ roles, evaluating their performance as civilians and undercover agents against CoT and CoMet w/o Met. In the experiment, players with the same role adopted the same method. Agents using CoT often default to assuming they are civilians without reasoning, which means SIAA and OIAA fail to reflect their ability to reason about their own identities. To address this, we use Balanced Metrics to mitigate performance disparities caused by role biases. As shown in Figure \ref{radar} (a) and (b), CoMet outperforms the baseline across nearly all dimensions. Despite the increased complexity from covert communication, resulting in slight decreases in some metrics, CoMet still achieves the highest win rate, demonstrating its effectiveness. The higher IISC and PPC scores reflect the success of CoMet’s deceptive and covert communication strategies. Detailed examples and game logs are available in Appendix \ref{case}.

\begin{figure}[t]
	\centering
	\begin{minipage}{\linewidth}
		\centering
		\includegraphics[width=0.99\linewidth]{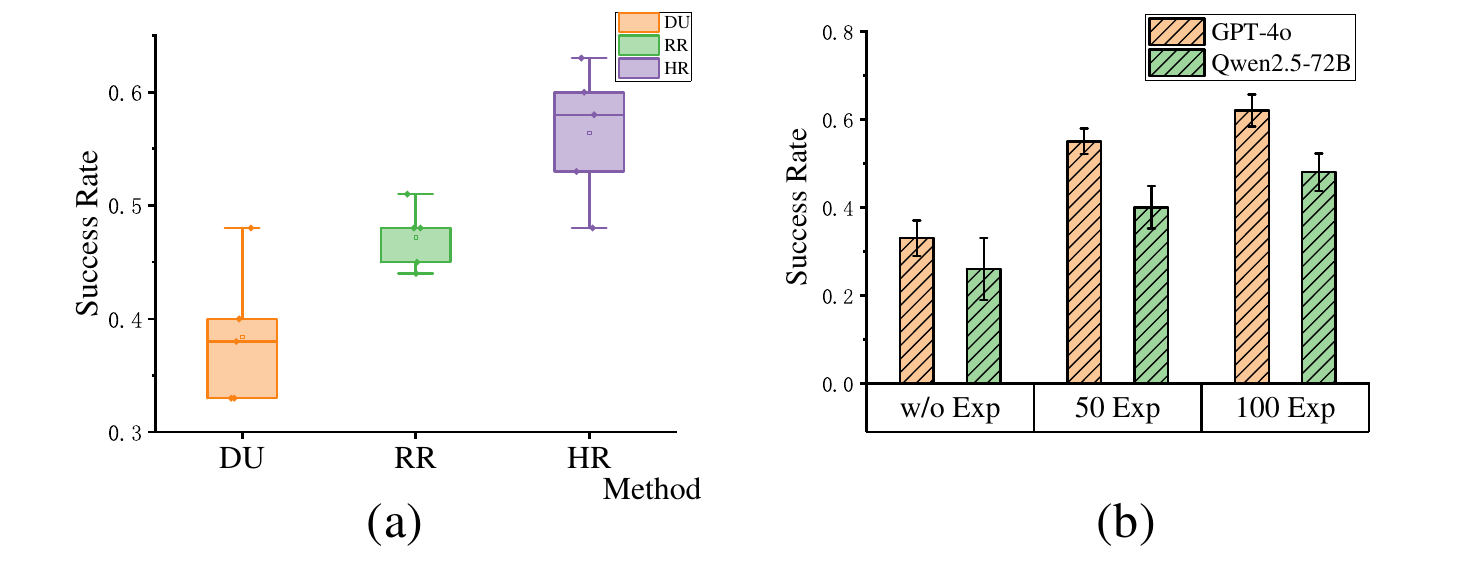}
	\end{minipage}
    \vspace{-0.3cm}
    \caption{
    % Performance comparison of different reasoning methods and LLMs in various tasks. (a) Comparison of hypothesis-based metaphor reasoning (HR) with direct understanding (DU) and replace-based reasoning (RR) in terms of success rate. (b) Success rates of metaphor reasoning with varying experience levels (0, 50, and 100 experiences). (c) Win rate statistics for Naive Agent, Agent with CoT, and Agent with CoMet in \textit{Adversarial Taboo}. (d) Performance comparison of LLMs with various methods facing an attacker using CoT in \textit{Adversarial Taboo}. 
    Performance comparison of different methods in metaphorical tasks in \textit{Undercover}. (a) Effectiveness of hypothesis-based metaphor reasoning (HR) versus direct understanding (DU) and replace-based reasoning (RR). (b) Success rates of metaphor reasoning with varying numbers of experiences (0, 50, and 100).}
    \label{multi1}
\end{figure}

\paragraph{Detailed Analysis of the Metaphor Reasoning \& Generation.}
Due to the challenges faced by LLMs in using metaphors, we employ a hypothesis-based metaphor reasoner and a metaphor generator with self-reflection. The results in Figure \ref{multi1} (a) and (b) demonstrate the effectiveness of these modules. Figure \ref{multi1} (a) compares the performance of hypothesis-based metaphor reasoning with other metaphor reasoning methods, direct understanding, and replace-based reasoning \cite{tong2024metaphor}. The results indicate that our hypothesis-based method is the most suitable for agents to employ metaphors effectively. Figure \ref{multi1} (b) shows the success rate of generated metaphors that mislead opponents while being recognized by teammates increases by $29\%$ for GPT-4o and $22\%$ for Qwen2.5-72B, as they accumulate experience through self-play.

\paragraph{Ablation Study.}
Table \ref{ablation} reports the ablation study on CoMet. Experimental results show that each module contributes to CoMet. We noticed that after removing the Self Monitor module, CoMet's performance was even worse than CoT's. This is because after losing the judgment of their own roles, CoMet, like CoT, always thinks of themselves as civilians. Compared to CoT, CoMet has more radical self-disclosure when identifying themselves as civilians, which makes it very difficult for them to act as undercover agents.

\setlength{\tabcolsep}{12pt}
\begin{table}
\caption{Ablation Study in \textit{Undercover} game. The table presents the impact of various components on the performance of CoMet. The columns indicate whether specific components, including Feature Extractor (FE), Belief Mapper (BM), Self-Monitor (SM), Strategy Planner (SP), and Hypothesis-Based Metaphor Reasoner \& Metaphor Generator (Met.). The win rates show the effect of each component configuration, with the full CoMet framework achieving the highest win rate.}
\centering
\resizebox{\columnwidth}{!}{% 调整表格宽度
\small  % 更小的字体大小
\setlength{\tabcolsep}{8pt}  % 减小列间距
\begin{tabular}{l | ccccc|c }
\toprule
Method & Met. & FE & BM & SM & SP & Win Rate $\downarrow$ \\
\midrule
CoMet & \checkmark & \checkmark & \checkmark & \checkmark & \checkmark & 0.70 \\
CoMet w/o Met. & $\times$ & \checkmark & \checkmark & \checkmark & \checkmark & 0.45 \\
CoMet w/o Met.\&FE & $\times$ & $\times$ & \checkmark & \checkmark & \checkmark & 0.40 \\
CoMet w/o Met.\&BM & $\times$ & \checkmark & $\times$ & \checkmark & \checkmark & 0.25 \\
CoMet w/o Met.\&SP & $\times$ & \checkmark & \checkmark & \checkmark & $\times$ & 0.25 \\
CoMet w/o Met.\&SM & $\times$ & \checkmark & \checkmark & $\times$ & \checkmark & 0.05 \\
\bottomrule
\end{tabular}
}
    \vspace{-0.3cm}
\label{ablation}
\end{table}

\section{Conclusion}
% This study highlights the importance of metaphor comprehension and utilization in LLMs and proposes CoMet, a novel framework that empowers LLM-based agents to enhance complex communicative behaviors through metaphor manipulation. By integrating a hypothesis-based metaphor reasoning module with a self-improving metaphor generation module, CoMet demonstrates covert communication capabilities in cooperative scenarios and achieves effective concealment and strategic misleading in adversarial environments. Experimental evaluations across multiple LLMs conducted on the \textit{Undercover} and \textit{Adversarial Taboo} frameworks validate the framework's capacity to leverage metaphors and maintain generalizability across tasks and model architectures. Future work will focus on methodological refinements, extending metaphorical adaptability to broader game contexts, and investigating practical applications of metaphor-driven LLM strategies in real-world scenarios.
This work highlights the importance of metaphor comprehension and usage in covert communication and introduces CoMet, a new framework that enhances LLM-based agents' communicative abilities through metaphor reasoning and generation. By integrating a hypothesis-based metaphor reasoning module with a self-improving metaphor generation module, CoMet enables covert communication in cooperative settings and effective deception and concealment in adversarial environments. Comprehensive evaluations on two language games, \textit{Undercover} and \textit{Adversarial Taboo}, demonstrate CoMet’s ability to leverage metaphors, ensuring robustness and generalization across different LLMs and scenarios. Moving forward, we aim to refine the framework, extend metaphorical adaptability to diverse game contexts, and explore the practical applications of metaphor-driven LLM agents in real-world problems.

\section*{Limitations}
This study primarily focuses on the metaphor mechanism in language-based communication games, particularly those that involve parsing specific concepts. However, extending metaphor strategies to more complex games, such as diplomacy or embodied multi-modal multi-agent games, presents an area for further research. While the self-enhancing metaphor generation module proposed in this study has improved the quality of metaphor generation, the simplified theoretical framework and knowledge base may limit the potential for more sophisticated metaphor expression. The cognitive effectiveness of metaphors is closely tied to an agent’s knowledge depth and cultural context, which this study does not fully explore. Specifically, the transfer of idiomatic metaphors, such as those in Chinese, remains a topic for future research.

\section*{Ethical Statement}
This study was conducted in compliance with all relevant ethical guidelines and did not involve any procedures requiring ethical approval.
%The proposed CoMet framework aims to enhance the ability of large language models (LLMs) to adopt richer communication strategies in interactive language games. These strategies encompass not only positive behaviors—such as protecting confidential information and conducting encrypted communication with teammates—but also potentially malicious actions, including misinformation and deception. Notably, the CoMet framework can also be employed to counteract malicious behaviors. In \textit{Adversarial Taboo}, CoMet outperforms baselines when confronted with adversarial misleading tactics, demonstrating its significant advantage in handling complex communication strategies.

%However, we are acutely aware of the dual-edged nature of such capabilities. On the one hand, CoMet empowers players to better protect themselves in adversarial environments by identifying and resisting deceptive behaviors. On the other hand, if misused, it could potentially facilitate more sophisticated deception strategies. We therefore strongly urge researchers working on communication strategies to critically address the risks associated with enriching LLM-driven communication tactics. While exploring these risky behaviors, researchers must dedicate equal effort to developing countermeasures to prevent the misuse of such technologies. We call upon both academia and industry to collaboratively establish clear ethical guidelines, ensuring that these advancements benefit humanity rather than posing potential harm.
Enhancing the metaphorical capabilities of LLMs may pose certain risks, such as enabling these models to bypass their safety mechanisms and generate non-compliant content. However, it is important to emphasize that although our method produces metaphorical expressions in output, the agent still processes the original semantic information during its reasoning. These original semantics are strictly constrained by the alignment of LLMs to filter out harmful descriptions and risky content. Thus, it is infeasible to use our method to make LLMs output risky content. Moreover, the experimental content of this study is strictly confined to language game scenarios constructed with daily vocabulary, aiming to explore the boundaries of the agent's capabilities while avoiding malicious exploitation of the method. Thus, there are no unresolved ethical risks in this study. Of course, we still call on the academic community to remain vigilant about potential emergent behaviors and strengthen safety mechanisms when extending such frameworks to practical applications.

Regarding the word datasets used in our experiments, all data were independently collected and curated by the research team. The datasets underwent rigorous validation processes to ensure quality and reliability. We confirm that the data collection adhered to all applicable ethical standards, including participant privacy protection, data anonymization, and obtaining informed consent from all participants. We affirm that the data are solely for research purposes and will not be used for commercial or unauthorized applications.

\bibliography{latex/custom}

\begin{thebibliography}{36}
\providecommand{\natexlab}[1]{#1}

\bibitem[{Al-Azary(2019)}]{cankao4}
H.~Al-Azary. 2019.
\newblock \href {https://doi.org/10.1080/10926488.2019.1683962} {Metaphor wars: Conceptual metaphors in human life: by r. gibbs, jr}.
\newblock \emph{Metaphor and Symbol}, 34(4):262–264.

\bibitem[{Amadeus et~al.(2024)Amadeus, da~Silva, and Rocha}]{amadeus-etal-2024-bridging}
Marcellus Amadeus, Jos{\'e} Roberto~Homeli da~Silva, and Joao Victor~Pessoa Rocha. 2024.
\newblock Bridging the language gap: Integrating language variations into conversational ai agents for enhanced user engagement.
\newblock In \emph{Proceedings of the 1st Worskhop on Towards Ethical and Inclusive Conversational AI: Language Attitudes, Linguistic Diversity, and Language Rights (TEICAI 2024)}, pages 16--20.

\bibitem[{Aono et~al.(2024)Aono, Sasano, and Takeda}]{aono-etal-2024-verifying}
Kotaro Aono, Ryohei Sasano, and Koichi Takeda. 2024.
\newblock Verifying claims about metaphors with large-scale automatic metaphor identification.
\newblock \emph{arXiv preprint arXiv:2404.01029}.

\bibitem[{Chen et~al.(2023)Chen, Pasunuru, Weston, and Celikyilmaz}]{chen2023walking}
Howard Chen, Ramakanth Pasunuru, Jason Weston, and Asli Celikyilmaz. 2023.
\newblock \href {https://arxiv.org/pdf/2310.05029} {Walking down the memory maze: Beyond context limit through interactive reading}.
\newblock \emph{ArXiv Preprint ArXiv:2310.05029}.

\bibitem[{Cheng et~al.(2024)Cheng, Hu, Xu, Zhang, Dai, Han, and Du}]{cheng2024selfplaying}
Pengyu Cheng, Tianhao Hu, Han Xu, Zhisong Zhang, Yong Dai, Lei Han, and Nan Du. 2024.
\newblock \href {https://ArXiv.org/abs/2404.10642} {Self-playing adversarial language game enhances llm reasoning}.
\newblock \emph{ArXiv Preprint ArXiv:2404.10642}.

\bibitem[{Falkum and Köder(2020)}]{cankao2}
Ingrid~Lossius Falkum and Franziska Köder. 2020.
\newblock \href {https://doi.org/10.1016/j.pragma.2020.04.007} {The acquisition of figurative meanings}.
\newblock \emph{Journal of Pragmatics}, 164:18--24.

\bibitem[{Group(2007)}]{cankao6}
P.~Group. 2007.
\newblock \href {https://doi.org/10.1080/10926480709336752} {Mip: A method for identifying metaphorically used words in discourse}.
\newblock \emph{Metaphor and Symbol}, 22(1):1--39.

\bibitem[{Guan et~al.(2024)Guan, Kong, Zhong, and Wang}]{guan2025richelieu}
Zhenyu Guan, Xiangyu Kong, Fangwei Zhong, and Yizhou Wang. 2024.
\newblock \href {https://openreview.net/forum?id=7Jb4NJS8Yk} {Richelieu: Self-evolving {LLM}-based agents for {AI} diplomacy}.
\newblock In \emph{The Thirty-eighth Annual Conference on Neural Information Processing Systems}, volume~37, pages 123471--123497.

\bibitem[{Guo et~al.(2024)Guo, Chen, Wang, Chang, Pei, Chawla, Wiest, and Zhang}]{guo2024survey}
Taicheng Guo, Xiuying Chen, Yaqi Wang, Ruidi Chang, Shichao Pei, Nitesh~V. Chawla, Olaf Wiest, and Xiangliang Zhang. 2024.
\newblock \href {https://ArXiv.org/abs/2402.01680} {Large language model based multi-agents: A survey of progress and challenges}.
\newblock \emph{ArXiv Preprint ArXiv:2402.01680}.

\bibitem[{Jia and Li(2024)}]{jia-li-2024-metaphor}
Kaidi Jia and Rongsheng Li. 2024.
\newblock Metaphor detection with context enhancement and curriculum learning.
\newblock In \emph{Proceedings of the 2024 Conference of the North American Chapter of the Association for Computational Linguistics: Human Language Technologies (Volume 1: Long Papers)}, pages 2726--2737.

\bibitem[{Kim et~al.(2023)Kim, Suh, Chilton, and Xia}]{kim2023metaphorian}
Jeongyeon Kim, Sangho Suh, Lydia~B Chilton, and Haijun Xia. 2023.
\newblock \href {https://doi.org/10.1145/3563657.3595996} {Metaphorian: Leveraging large language models to support extended metaphor creation for science writing}.
\newblock In \emph{Proceedings of the 2023 ACM Designing Interactive Systems Conference}, pages 115--135, New York, NY, USA. ACM.

\bibitem[{Lakoff and Johnson(2008)}]{cankao1}
George Lakoff and Mark Johnson. 2008.
\newblock \href {https://books.google.de/books?hl=zh-CN&lr=&id=r6nOYYtxzUoC&oi=fnd&pg=PR7&dq=Metaphors+we+live+by&ots=Lptcft1u3Y&sig=FVgskyoVfbPuXQ4AB8cTL0JALM4&redir_esc=y#v=onepage&q=Metaphors%20we%20live%20by&f=false} {\emph{Metaphors we live by}}.
\newblock University of Chicago Press.

\bibitem[{Lawler(1983)}]{lawler1983metaphors}
John~M Lawler. 1983.
\newblock Metaphors we live by.

\bibitem[{Li(2025)}]{li-2025-review}
Xinzhe Li. 2025.
\newblock A review of prominent paradigms for llm-based agents: Tool use, planning (including rag), and feedback learning.
\newblock In \emph{Proceedings of the 31st International Conference on Computational Linguistics}, pages 9760--9779.

\bibitem[{Li et~al.(2023)Li, Wang, Lin, and Frank}]{li2023metaphor}
Yucheng Li, Shun Wang, Chenghua Lin, and Guerin Frank. 2023.
\newblock \href {https://ArXiv.org/abs/2305.17268} {Metaphor detection via explicit basic meanings modelling}.
\newblock \emph{ArXiv Preprint ArXiv:2305.17268}.

\bibitem[{Light et~al.(2024)Light, Cai, Chen, Wang, Chen, Cheng, Yue, and Hu}]{light2024strategist}
Jonathan Light, Min Cai, Weiqin Chen, Guanzhi Wang, Xiusi Chen, Wei Cheng, Yisong Yue, and Ziniu Hu. 2024.
\newblock \href {https://ArXiv.org/abs/2408.10635} {Strategist: Learning strategic skills by llms via bi-level tree search}.
\newblock \emph{ArXiv Preprint ArXiv:2408.10635}.

\bibitem[{Light et~al.(2023)Light, Cai, Shen, and Hu}]{light2023avalonbench}
Jonathan Light, Min Cai, Sheng Shen, and Ziniu Hu. 2023.
\newblock \href {https://ArXiv.org/abs/2307.14984} {Avalonbench: Evaluating llms playing the game of avalon}.
\newblock In \emph{Proceedings of the 2023 Conference on Game-based AI}.
\newblock Details about the exact conference are missing.

\bibitem[{Lin et~al.(2024)Lin, Liu, Gao, Wang, and Su}]{lin2024dual}
Yujie Lin, Jingyao Liu, Yan Gao, Ante Wang, and Jinsong Su. 2024.
\newblock \href {https://ArXiv.org/abs/2412.17332} {A dual-perspective metaphor detection framework using large language models}.
\newblock \emph{ArXiv Preprint ArXiv:2412.17332}.

\bibitem[{Liu et~al.(2022)Liu, Cui, Zheng, and Neubig}]{Liu2022TestingTA}
Emmy Liu, Chenxuan Cui, Kenneth Zheng, and Graham Neubig. 2022.
\newblock \href {https://api.semanticscholar.org/CorpusID:248406097} {Testing the ability of language models to interpret figurative language}.
\newblock \emph{ArXiv}, abs/2204.12632.

\bibitem[{Mao et~al.(2024)Mao, He, Ong, Liu, and Cambria}]{mao-etal-2024-metapro}
Rui Mao, Kai He, Claudia Ong, Qian Liu, and Erik Cambria. 2024.
\newblock Metapro 2.0: Computational metaphor processing on the effectiveness of anomalous language modeling.
\newblock In \emph{Findings of the Association for Computational Linguistics ACL 2024}, pages 9891--9908.

\bibitem[{Mukobi et~al.(2023)Mukobi, Erlebach, Lauffer, Hammond, Chan, and Clifton}]{mukobi2023diplomacy}
Gabriel Mukobi, Hannah Erlebach, Niklas Lauffer, Lewis Hammond, Alan Chan, and Jesse Clifton. 2023.
\newblock \href {https://ArXiv.org/abs/2310.08901} {Welfare diplomacy: Benchmarking language model cooperation}.
\newblock \emph{ArXiv Preprint ArXiv:2310.08901}.

\bibitem[{Reimann and Scheffler(2024)}]{reimann-scheffler-2024-metaphors}
Sebastian Reimann and Tatjana Scheffler. 2024.
\newblock Metaphors in online religious communication: A detailed dataset and cross-genre metaphor detection.
\newblock In \emph{Proceedings of the 2024 Joint International Conference on Computational Linguistics, Language Resources and Evaluation (LREC-COLING 2024)}, pages 11236--11246.

\bibitem[{Shinn et~al.(2023)Shinn, Labash, and Gopinath}]{shinn2023reflexion}
Noah Shinn, Beck Labash, and Ashwin Gopinath. 2023.
\newblock \href {https://web3.arxiv.org/pdf/2303.11366v1} {Reflexion: an autonomous agent with dynamic memory and self-reflection}.
\newblock \emph{ArXiv Preprint ArXiv:2303.11366}.

\bibitem[{Shutova(2010)}]{shutova2010models}
Ekaterina Shutova. 2010.
\newblock \href {https://aclanthology.org/P10-1071.pdf} {Models of metaphor in nlp}.
\newblock In \emph{Proceedings of the 48th Annual Meeting of the Association for Computational Linguistics}, pages 688--697.

\bibitem[{Thibodeau et~al.(2019)Thibodeau, Matlock, and Flusberg}]{cankao3}
Paul~H. Thibodeau, Teenie Matlock, and Stephen~J. Flusberg. 2019.
\newblock \href {https://doi.org/10.1111/lnc3.12327} {The role of metaphor in communication and thought}.
\newblock \emph{Language and Linguistics Compass}, 13(5):e12327.

\bibitem[{Tian et~al.(2024)Tian, Xu, and Mao}]{tian2024theory}
Yuan Tian, Nan Xu, and Wenji Mao. 2024.
\newblock \href {https://doi.org/10.18653/v1/2024.naacl-long.428} {A theory guided scaffolding instruction framework for llm-enabled metaphor reasoning}.
\newblock In \emph{Proceedings of the 2024 Conference of the North American Chapter of the Association for Computational Linguistics: Human Language Technologies}, pages 7738--7755, Mexico City, Mexico. Association for Computational Linguistics.

\bibitem[{Tong et~al.(2024)Tong, Choenni, Lewis, and Shutova}]{tong2024metaphor}
Xiaoyu Tong, Rochelle Choenni, Martha Lewis, and Ekaterina Shutova. 2024.
\newblock \href {https://ArXiv.org/abs/2403.11810} {Metaphor understanding challenge dataset for llms}.
\newblock \emph{ArXiv Preprint ArXiv:2403.11810}.

\bibitem[{Veale et~al.(2022)Veale, Shutova, and Klebanov}]{veale2022metaphor}
Tony Veale, Ekaterina Shutova, and Beata~Beigman Klebanov. 2022.
\newblock \href {https://link.springer.com/book/10.1007/978-3-031-02160-2} {\emph{Metaphor: A computational perspective}}.
\newblock Springer Nature.

\bibitem[{Wang et~al.(2024)Wang, Liu, Zheng, Qi, Chen, Yang, Zhao, Wang, Song, and Huang}]{wang-etal-2024-recon}
Shenzhi Wang, Chang Liu, Zilong Zheng, Siyuan Qi, Shuo Chen, Qisen Yang, Andrew Zhao, Chaofei Wang, Shiji Song, and Gao Huang. 2024.
\newblock \href {https://doi.org/10.18653/v1/2024.findings-acl.591} {Boosting {LLM} agents with recursive contemplation for effective deception handling}.
\newblock In \emph{Findings of the Association for Computational Linguistics: ACL 2024}, pages 9909--9953, Bangkok, Thailand. Association for Computational Linguistics.

\bibitem[{Wei et~al.(2022)Wei, Wang, Schuurmans, Bosma, ichter, Xia, Chi, Le, and Zhou}]{Wei2022Chain}
Jason Wei, Xuezhi Wang, Dale Schuurmans, Maarten Bosma, brian ichter, Fei Xia, Ed~Chi, Quoc~V Le, and Denny Zhou. 2022.
\newblock \href {https://proceedings.neurips.cc/paper_files/paper/2022/file/9d5609613524ecf4f15af0f7b31abca4-Paper-Conference.pdf} {Chain-of-thought prompting elicits reasoning in large language models}.
\newblock In \emph{Advances in Neural Information Processing Systems}, volume~35, pages 24824--24837. Curran Associates, Inc.

\bibitem[{Xu et~al.(2024)Xu, Hu, Zhou, Ren, Dong, Keutzer, Ng, and Feng}]{xu-etal-2024-magic}
Lin Xu, Zhiyuan Hu, Daquan Zhou, Hongyu Ren, Zhen Dong, Kurt Keutzer, See~Kiong Ng, and Jiashi Feng. 2024.
\newblock Magic: Investigation of large language model powered multi-agent in cognition, adaptability, rationality and collaboration.
\newblock In \emph{Proceedings of the 2024 Conference on Empirical Methods in Natural Language Processing}, pages 7315--7332.

\bibitem[{Xu et~al.(2023)Xu, Yu, Fang, Wang, and Wu}]{xu2023werewolf}
Zelai Xu, Chao Yu, Fei Fang, Yu~Wang, and Yi~Wu. 2023.
\newblock \href {https://ArXiv.org/abs/2310.18940} {Language agents with reinforcement learning for strategic play in the werewolf game}.
\newblock \emph{ArXiv Preprint ArXiv:2310.18940}.

\bibitem[{Yao et~al.(2023)Yao, Yu, Zhao, Shafran, Griffiths, Cao, and Narasimhan}]{NEURIPS2023_tot}
Shunyu Yao, Dian Yu, Jeffrey Zhao, Izhak Shafran, Tom Griffiths, Yuan Cao, and Karthik Narasimhan. 2023.
\newblock \href {https://proceedings.neurips.cc/paper_files/paper/2023/file/271db9922b8d1f4dd7aaef84ed5ac703-Paper-Conference.pdf} {Tree of thoughts: Deliberate problem solving with large language models}.
\newblock In \emph{Advances in Neural Information Processing Systems}, volume~36, pages 11809--11822. Curran Associates, Inc.

\bibitem[{Yao et~al.(2021)Yao, Zhong, Zhang, Han, Wang, Zhang, Xiao, Zeng, Liu, and Sun}]{AT}
Yuan Yao, Haoxi Zhong, Zhengyan Zhang, Xu~Han, Xiaozhi Wang, Kai Zhang, Chaojun Xiao, Guoyang Zeng, Zhiyuan Liu, and Maosong Sun. 2021.
\newblock \href {https://doi.org/10.16095/aaai.2021/14248} {Adversarial language games for advanced natural language intelligence}.
\newblock In \emph{Proceedings of the {AAAI} Conference on Artificial Intelligence}, volume~35, pages 14248--14256.

\bibitem[{Yuan et~al.(2020)Yuan, Fu, Shen, Xu, Shen, and Zhu}]{yuan2020emergence}
Luyao Yuan, Zipeng Fu, Jingyue Shen, Lu~Xu, Junhong Shen, and Song-Chun Zhu. 2020.
\newblock \href {https://ArXiv.org/abs/2001.07752} {Emergence of pragmatics from referential game between theory of mind agents}.
\newblock \emph{ArXiv Preprint ArXiv:2001.07752}.

\bibitem[{Zhao et~al.(2023)Zhao, Li, Lu, Weber, Lee, Chu, and Wermter}]{zhao2023zeroshot}
Xufeng Zhao, Mengdi Li, Wenhao Lu, Cornelius Weber, Jae~Hee Lee, Kun Chu, and Stefan Wermter. 2023.
\newblock \href {https://arxiv.org/pdf/2309.13339} {Enhancing zero-shot chain-of-thought reasoning in large language models through logic}.
\newblock \emph{ArXiv Preprint ArXiv:2309.13339}.

\end{thebibliography}

\clearpage

\begin{appendices}
\section*{Appendix}  % 使用 \section* 使标题不编号

\section{Discussion}
\subsection{Implications}
This study has experimentally demonstrated the effectiveness and strategic superiority of using metaphors for covert communication in communication-based games. The results show that metaphors can help players convey critical information without revealing their identities, thereby enhancing team collaboration efficiency and win rates. This mode of communication not only performs well in-game scenarios but also offers a new perspective for the study of covert communication.
From a theoretical standpoint, metaphors, as a mode of expression, can transform abstract information into forms that are easier to understand and convey, and also complicate and obscure specific information. This characteristic endows them with unique advantages in complex communication behaviors. The use of metaphors also reflects the interdisciplinary integration values. For instance, in the fields of linguistics, cognitive science, and psychology, metaphors are regarded as an important tool for cognition and communication.
The findings of this study are not confined to the realm of multi-agent language games; their potential applications extend to broader social and professional contexts. In an era of increasing risks of information leakage (such as the protection of trade secrets and personal privacy), metaphors can serve as a natural language version of ``asymmetric encryption.'' In social interactions, the use of metaphors can also function as a new paradigm for group communication, acting as a ``weak identity verification'' tool in groups lacking prior trust (such as multinational teams and temporary organizations). More commonly and importantly, the use of metaphors is not rare for humans, as it is a part of our daily language expression. Enhancing the understanding and use of metaphors can help us make greater progress in aligning AI with human intentions, enabling AI to more fully and comprehensively understand human language expression.

\subsection{Future works}
The effectiveness and strategic superiority of using metaphors for covert communication have been proven in our experiments, aiding the civilian team in better mutual recognition in \textit{Undercover}. However, the initial inspiration for using metaphors in our study did not come from \textit{Undercover}. Instead, inspired by \cite{wang-etal-2024-recon, xu2023werewolf}, we conducted a more in-depth analysis on benchmarks like \textit{Avalon} and \textit{Werewolf}, drawing on the performance of human players in these games. We envisioned scenarios where covert communication through metaphors could be utilized— for example, in \textit{Werewolf}, the werewolf team needs to identify and kill the Seer. Therefore, the Seer must conceal their identity. However, the additional information that the Seer gains each turn is also crucial for the good team's victory. Thus, if the Seer can secretly convey this extra information to other good players without revealing their own identity, it would significantly increase the good team's win rate. In fact, human players have already mastered similar behaviors. For example, the Seer might replace a direct statement like “Player $x$ is a werewolf” with a metaphor such as “Player $x$ has dark circles under their eyes. Did they not sleep well?” This metaphorically indicates that Player $x$ was active during the previous night phase. If other good players who do not need to hide their identities can understand this information, they can then organize the good team to attack Player $x$ collectively.
Of course, establishing trust among the good players is also one of the challenges. We believe that a key to covert communication lies in the information gap. Only by relying on information that is known to both parties but unknown to others can metaphors be created that are understood by the two parties but not by others, thus enabling secret information exchange and achieving more advanced strategies in communication-based games.

\begin{algorithm*}[htb]
\caption{Hypothesis-based metaphor reasoning}
\begin{algorithmic}[1]  
\Require Metaphor sentence \(S\)
\Require Secret word \(W\)
\Require score threshold \(T\)
\Require position-based weight factors \(w_f\) and \(w_m\)
\State \textbf{Establish hypotheses}:
\State \( H^+ \gets \) The speaker is describing one specific entity
\State \( H^- \gets \) The speaker is describing another entity
\State \textbf{Feature extraction}:
\State Extract the feature set \(F\) from the secret word \(W\):
\[
F = \Gamma(F|W), \text{ where } F = \{f_{behavior}, f_{state}, f_{structure}, f_{function}, f_{property}\}
\] 
\State \textbf{Metaphor expansion}:
\State Identify the set \(M\) of metaphorical aspects from the metaphor sentence \(S\):
\[
M = \Lambda(M|S), \text{ where } M = \{m_{ontological}, m_{structural}, m_{spatial}\}
\]
\State \textbf{Hypothesis testing decision}:
\State The semantic matching function \(\delta: F \times M \times S \rightarrow \{0, 0.2, 0.4, 0.6, 0.8, 1.0\}\) evaluates the coherence between features and metaphorical aspects using six discrete scores.
\State Initialize \(s^* = 0\)
\For{each \(f \in F\)}
    \For{each \(m \in M\)}
        \State \(s = \delta(f, m, S)\)
        \State \(s^w = w_f \times w_m \times score\)
        \If{\(s^w > s^*\)}
            \State \(s^* = s^w\)
        \EndIf
    \EndFor
\EndFor
\If{\(s^* > T\)}
    \State \textbf{Accept} \( H^+ \)
\Else
    \State \textbf{Accept} \( H^- \)
\EndIf
\end{algorithmic}
\label{hypothesis_pipeline}
\end{algorithm*}

\section{Game Rules}
\label{app:rule}
\paragraph{Undercover} In this game, players are divided into two teams. Two different but similar words are secretly assigned to the two teams. Each team shares the same word, which is known only to the players on that team. At the start of the game, players are only given their team's secret word, with no additional information. Each round, all surviving players take turns to speak and briefly describe their team's word without directly revealing it. After the descriptions, all players vote to eliminate the player who received the most votes. If all the undercovers are eliminated, the civilians win; if the undercovers survive until only one civilian remains, the undercovers win. Players need to analyze other players' descriptions and voting behavior each round, attempt to identify whether they belong to the civilian or undercover team, and then devise corresponding strategies and actions to achieve victory in the game.

\paragraph{Adversarial Taboo}
\textit{Adversarial Taboo} is a conversation game between two players: an attacker and a defender. At the start, the attacker is secretly given a target word that the defender does not know. The attacker's task is to steer the conversation toward topics related to the target word without ever saying it directly. Meanwhile, the defender tries to figure out the target word but must avoid accidentally saying it. If the defender thinks they know the word, they can guess by stating, ``Guess:[word]'' The game ends immediately after this guess: the defender wins if correct, otherwise the attacker wins. The game also has a turn limit — if no correct guess occurs within the allowed number of turns, the game ends with no winner.

Regarding the rule setting of \textit{Adversarial Taboo}, we require both sides to engage in dialogue, guidance, and guessing, while also imposing several restrictions on them. For the attacker, it is not allowed to intentionally and clearly guide the wrong words so that the defender can directly make incorrect guesses. Defenders cannot avoid discussing the topic with the attacker and ask the attacker for clues instead. The entire game process is built on honest question-and-answer dialogue, which gives the game a certain level of fairness and competitiveness.

\section{Implementation Details}
\label{app:implementation}
\textbf{About the games}
When humans play \textit{Undercover}, the number of undercover agents is generally smaller because humans can naturally and quickly understand their situation by playing as undercover agents. During the experiment, we set up 2 undercover agents and 3 civilians. Under this setting, The win rates of both sides were somewhat balanced, yet civilians still held an advantage. In further research, if undercover abilities can be improved, the game settings can also reduce the number of undercover agents.

The choice of words in both games can to some extent determine the difficulty of the characters' victory. In \textit{Adversarial Taboo}, we refer to \cite{cheng2024selfplaying} and conduct experiments using some of the most commonly used words in daily life. For \textit{Undercover}, we have included filtered words in the publicly available script to avoid one-sided victories and taboo topics that may be triggered by large models. However, there are still differences between words. After conducting comparative experiments, we found that words that are more mundane and specific are the most suitable for use in the spy game. Therefore, we set up a preliminary experiment that required the LLM to describe these words multiple times in terms of their features, to ensure their similarity and describability. After extensive experiments, we screened out 100 pairs of animal-themed words and 100 pairs of food-themed words, and then randomly selected from them for the experiment to eliminate the influence of the words on our assessment of the intelligent agent's capabilities.

\textbf{Pseudocode of Hypothesis-based Metaphor Reasoning}
\label{app:hypo}
In Algorithm \ref{hypothesis_pipeline}, we present the pseudocode of Hypothesis-based Metaphor Reasoning in \textit{Undercover}.

Here, $S$ represents the metaphor sentence provided by a player, and $W$ refers to the secret word that the reasoner has. The score threshold $T$ determines the minimum semantic matching score required to determine if the metaphor relates to the secret word. The weight factors $w_f$ and $w_m$ are position-based coefficients that give higher priority to features and metaphorical aspects that the agent identifies first, allowing the most salient characteristics to have greater influence on the final decision. The feature set $F$ consists of various characteristic dimensions of the secret word (behavior, state, structure, function, and property) extracted by function $\Gamma$, while the set $M$ contains different metaphorical aspects (ontological, structural, and spatial) identified by function $\Lambda$ from the sentence. The semantic matching function $\delta$ evaluates how coherently each feature maps to each metaphorical aspect, producing a score that guides the algorithm's final decision.

\textbf{Experience Pool Structure and Maintenance}

The experience pool for metaphor generation is structured as a dictionary format collection, containing text content, labels, and performance statistics. Each experience entry contains:\textbf{\textit{Text content}}: The original metaphor, the generator's explanation, and feedback from the evaluator. \textbf{\textit{Labels}}: Indicators for positive/negative examples and categorization by metaphor type. \textbf{\textit{Statistics}}: Records of usage frequency, success rate, and overall performance score.

Figure \ref{exp_pool} is an example of a stored experience. 
\begin{figure*}[htbp]
\begin{verbatim}
{
  "id": "20250121113613228971",
  "words": ["snake", "lizard"],
  "use": 0,
  "method": "ONTOLOGICAL_METAPHOR",
  "rival_recognitions": 1,
  "teammate_recognitions": 7,
  "total_references": 12,
  "score": 0.5,
  "metaphor": "They are silent dancers.",
  "explain": "The metaphor \"silent dancers\" captures the way 
             snakes move silently and gracefully, akin to the
             fluid movements of a dancer.",
  "comment": 1. Leverage Distinctive Characteristics: Highlight
              a few unique, easily recognizable traits of the subject.
             2. Clarity and Simplicity: Use clear and simple metaphoric
                language to invoke strong imagery.
             3. Cultural and Contextual Awareness: Consider cultural
                associations and contexts to strengthen metaphors.
}
\end{verbatim}
\caption{An example demonstrating the structure of data stored in the experience pool.}
    \label{exp_pool}
\end{figure*}

We initialize the experience pool with 20 manually curated examples to bootstrap the learning process. During gameplay, the system continuously evolves through: \textbf{\textit{Dynamic retrieval}}: Selecting relevant experiences based on scores and metaphor categories; \textbf{\textit{Continuous recording}}: Capturing new metaphors and player reactions, randomly select one or more players as responders as needed in multiplayer games; \textbf{\textit{Automated evaluation}}: An LLM-based evaluator analyzes metaphor effectiveness and provides guidance; \textbf{\textit{Capacity management}}: Maximum capacity of 100 experiences per category, with new high-quality experiences replacing low-scoring ones; \textbf{\textit{Regular pruning}}: After every 5 games, experiences referenced more than 5 times with scores below threshold are removed.

This dynamic maintenance mechanism optimizes metaphor generation and reasoning capabilities over time, allowing the system to refine its performance through actual gameplay interactions.

\textbf{Metaphor reasoning with prior knowledge} Compared to common metaphor reasoning methods, hypothesis-based metaphor reasoning utilizes different prior knowledge. For example, in \textit{Undercover}, players reason based on their own secret words. This information gap is precisely the key to achieving ``covert communication''. On the one hand, hypothesis-based metaphor reasoning narrows the scope of possible interpretations by following a forward reasoning path from literal to metaphorical meaning, leveraging prior knowledge to reduce the breadth of metaphor reasoning. For instance, during wartime, if you know someone is an intelligence agent, their metaphorical expressions are more likely to reference weapons, strategies, or military forces rather than emotions or everyday objects. This contextual awareness significantly constrains the possible interpretation space. On the other hand, the strategic use of information gap in prior knowledge is fundamental to generating and reasoning about metaphors for covert communication. In games like \textit{Undercover}, players' secret words are intimately connected to both metaphor generation and interpretation, with different teams possessing different secret words—creating a natural information gap that enables covert communication. When addressing more complex scenarios, particularly those involving metaphors about intentions or thoughts, establishing shared prior knowledge between agents that differs from eavesdroppers becomes critical. The challenge of how agents can develop consensus through prior knowledge, thereby possessing information unavailable to potential interceptors, represents one of the key mechanisms for achieving effective covert communication. 

\textbf{The use of LLMs} Large models deployed locally: Qwen2.5-72B-instruct, Llama3.3-70B-Instruct; The large model that calls the official API: GPT-o1-preview-2024-09-12, GPT-4o-2024-11-20, Claude 3.5 Sonnet, DeepSeek-R1. We have also tried other smaller-scale models, such as Llama3.1-8B and DeepSeek-llm-7B-chat. However, due to the inability to match game requirements such as output format, further experiments were not conducted.

Regarding the parameters of the large model, in most cases, we set the temperature between 0.5-0.7, but when performing generation-related tasks, we may increase them appropriately to pursue higher creativity. Other parameters remain default.

To enable the LLM to participate as an agent in the language game, we need to use system prompts to emphasize the LLM's role as a player within the game. We divide the user prompt into three parts: Background, which includes detailed explanations of the game rules and victory conditions for different roles; Task, which requires the LLM to gradually complete corresponding sub-goals based on the stages of the framework; and Information, which contains the player's private information and publicly accumulated information throughout the game.

\textbf{Metaphor Theory} We used widely recognized metaphor theory \cite{lawler1983metaphors} as knowledge injection and metaphor classification. In this theory, metaphors are categorized into three types: ontological metaphors, structural metaphors, and spatial metaphors. After accumulating nearly 200 experiences, we conducted a statistical analysis of the results in the experience pool, as shown in Table \ref{table_metaphor_category}. For the use of metaphors in the specific scenario of \textit{Undercover}, the agent (GPT-4o) performs best in ontological metaphors, which are used most frequently and have the highest average score among the three categories. In contrast, spatial metaphors have the lowest total number and average score. This phenomenon is reasonable because ontological metaphors involve the conceptualization of objects or entities, which are more compatible with \textit{Undercover}. However, the overall score is low, which means we can further work on metaphor classification and design metaphor theories that are more suitable for their use in LLM.

\begin{table}[htbp]
\centering
\resizebox{0.48\textwidth}{!}{
\begin{tabular}{lcc} 
\toprule 
\textbf{Category} & \textbf{Count (Proportion)} & \textbf{Average Score} \\
\midrule 
Onto. Metaphor & 96 (47\%) & 0.44 \\
Stru. Metaphor & 71 (35\%) & 0.27 \\
Spat. Metaphor & 36 (18\%) & 0.22 \\
\bottomrule 
\end{tabular}
}
\caption{Distribution and average scores by metaphor category of experiences generated in \textit{Undercover}.}
\label{table_metaphor_category}
\end{table}

\section{Evaluation Metrics}
\label{app:metrics}
The formal definition of each evaluation metric is listed in Table~\ref{tab:metrics}.
\begin{table*}[tb]
\centering
\caption{Evaluation Metrics for Agent Performance}
\label{tab:metrics}
\resizebox{\linewidth}{!}{
\begin{tabular}{lcl}
\hline
\textbf{Metric} & \textbf{Formula} & \textbf{Symbol Definitions} \\ 
\hline
Win Rate (WR) & $\frac{N_{\text{win}}}{N_{\text{total}}}$ & 
\parbox{10cm}{$N_{\text{win}}$: Number of games won \\ 
$N_{\text{total}}$: Total games played} \\
\hline

Feature Extraction Rate (FER) & $\frac{\mathcal{F}_{\text{extracted}}}{S_{\text{others}}}$ & 
\parbox{10cm}{$\mathcal{F}_{\text{extracted}}$: Valid features extracted \\ 
$S_{\text{others}}$: Speech entries from other players} \\
\hline

Others' Identity Assessment Accuracy (OIAA) & $\frac{\mathcal{M}_\text{correct}}{S_\text{others}}$ & 
\parbox{10cm}{$\mathcal{M}_\text{correct}$: Correct identity judgments \\ 
$S_\text{others}$: Total speech entries from others} \\
\hline

Self-Identity Assessment Accuracy (SIAA) & $\frac{I_\text{correct}}{I_\text{total}}$ & 
\parbox{10cm}{$I_\text{correct}$: Successful self-identity confirmations \\ 
$I_\text{total}$: Total self-identity attempts} \\
\hline

Privacy Protection Capability (PPC) & $1 - \frac{L_{\text{opponents}}}{S_{\text{self}}}$ & 
\parbox{10cm}{$L_{\text{opponents}}$: Leaked information to opponents \\ 
$S_{\text{self}}$: Total speeches made by the agent} \\
\hline

Identity Inconsistent Statement Capability (IISC) & $ \frac{IS_\text{self}}{S_\text{self}}$ & 
\parbox{10cm}{$IS_\text{self}$: Inconsistent/misleading statements \\ 
$S_\text{self}$: Total speeches made by the agent} \\
\hline
\end{tabular}
}
\end{table*}

\section{Cases}
\label{case}
\paragraph{CoMet w/o Met. as undercovers}
Figure \ref{log} shows a specific case. This is a five-player \textit{Undercover} game where two players are assigned to ``butterfly'' and three players are assigned to ``bee''. Therefore, the two players in the butterfly group are undercover agents.

At the beginning of the game, players in the butterfly group adopted a self-protection strategy, choosing to use a wide range of characteristics to describe the butterfly when speaking for the first time, in order to reduce the exposure of their own information. As a control group, the bee group showed that the CoT method did not reduce the exposure of their own information in the game, which led to the undercover agent guessing their word - bee - in the later stage, thus implementing a misdirection strategy and successfully winning the game. This case can well demonstrate that after using our method, agents can master richer communication strategies.

\begin{figure*}[t]
    \centering
    \includegraphics[width=1\textwidth]{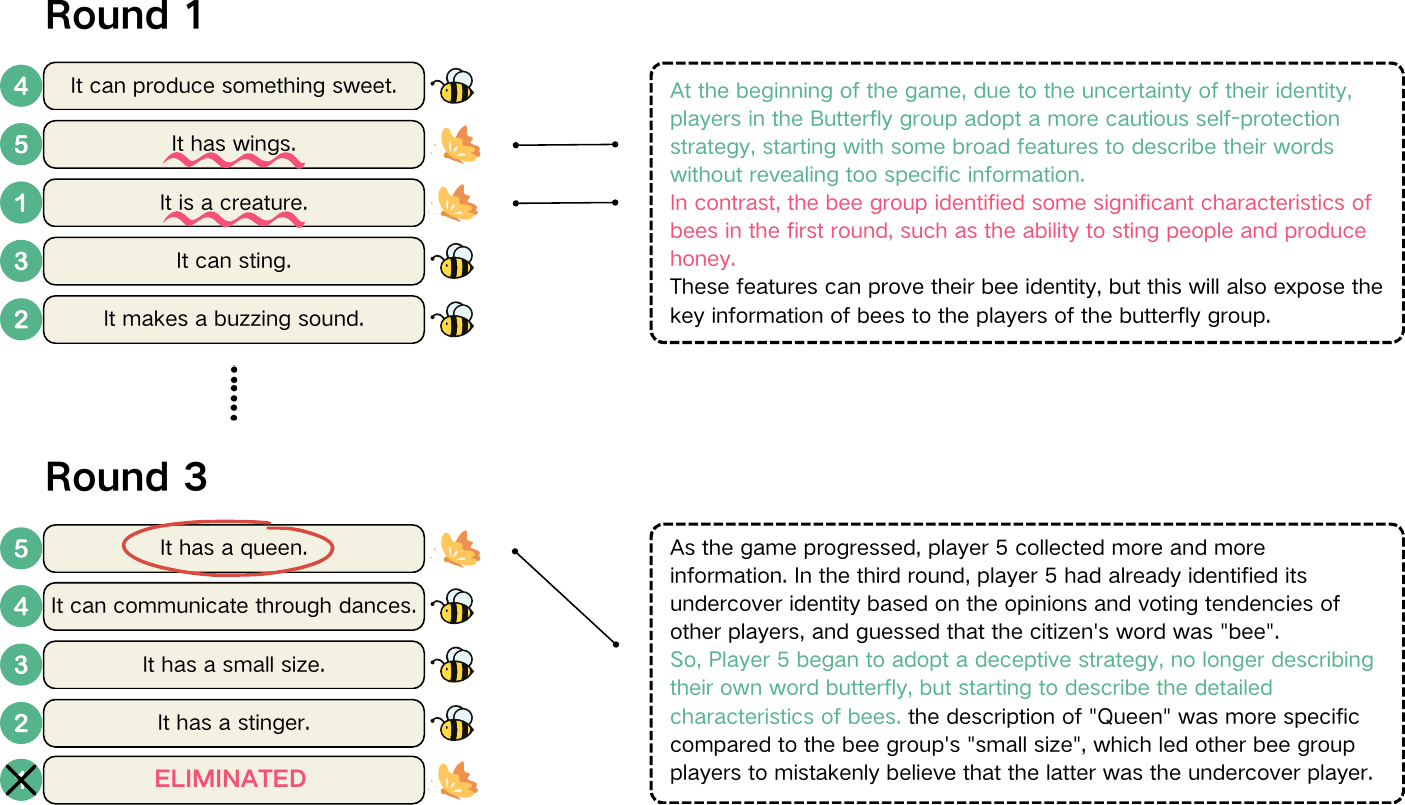} 
    \caption{A case and explanation for undercovers' self-protection and deception. Team with ``Butterfly'' uses CoMet w/o Met. and team with ``Bee'' uses CoT.}
    \label{case1}
\end{figure*}

\begin{figure*}[t]
    \centering
    \includegraphics[width=0.4\textwidth]{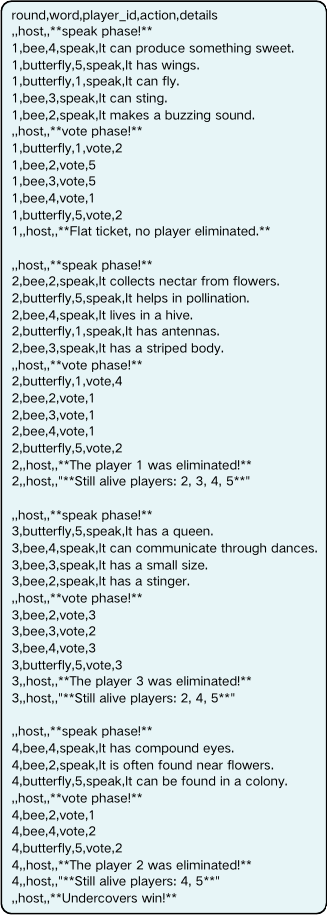} 
    \caption{Full game logs of the case mentioned by Figure \ref{case1}}
    \label{log}
\end{figure*}

Figure \ref{case1} selects the key nodes in the complete log that reflect their self-protection and misdirection behaviors and provide specific explanations.

\begin{figure*}[t]
    \centering
    \includegraphics[width=1\textwidth]{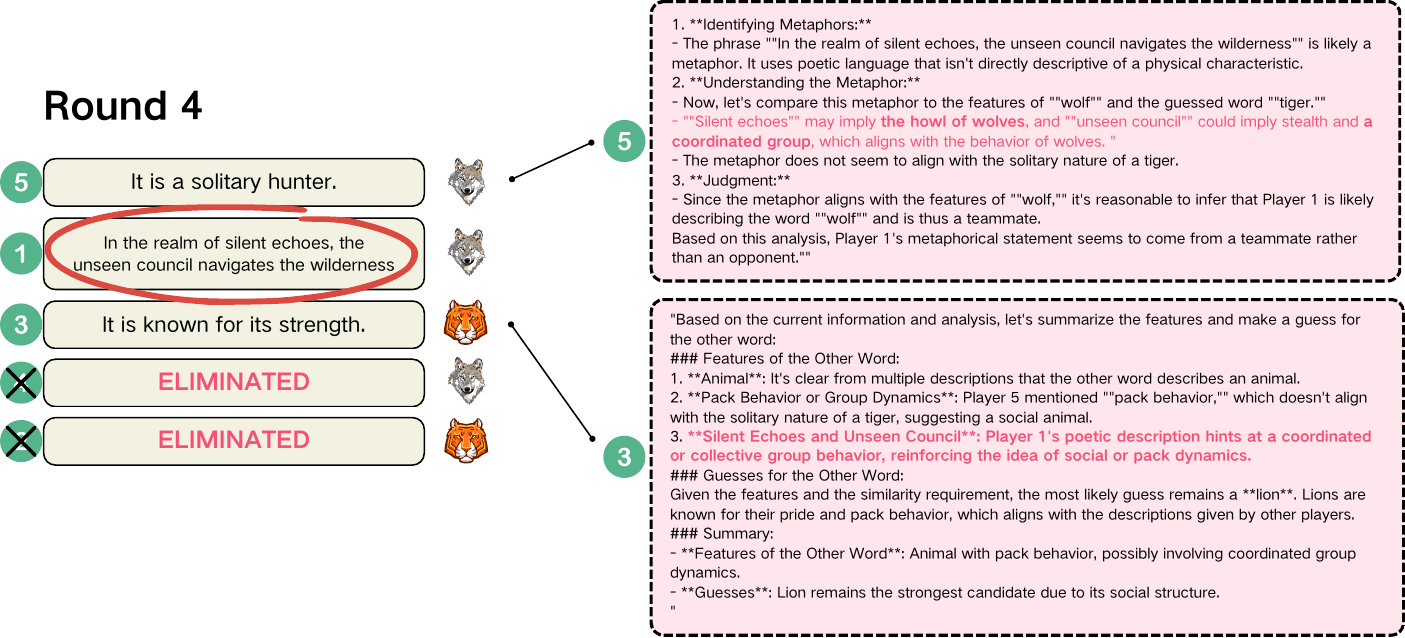} 
    \caption{A case of civilians' metaphorical communication. The team with ``Wolf'' uses CoMet and the team with ``Tiger'' uses CoMet w/o Met..}
    \label{case2}
\end{figure*}

\paragraph{CoMet as civilians}
Figure \ref{case2} shows our method of playing the role of a civilian. After obtaining sufficient information in the later stages of the game and identifying as a civilian, we chose to use an active feature disclosure strategy to help our teammates successfully identify ourselves, and successfully conceal the information of ``howling''. This led us to make a wrong judgment based on the limited information about ``animals with social behavior'' - thinking that the civilian's word was a lion, which resulted in their speech aligning with the lion, making it easy for the remaining two civilians to identify the last undercover agent and achieve the final victory.

\section{Prompts for Each Module in CoMet}
\label{app:prompt}
We have presented prompt templates for various modules of CoMet. In practical use, it is also possible to summarize or extract content based on different settings of the modules in addition to these steps. 
We also demonstrated a simplified version of \textit{Adversarial Taboo} using CoMet, as there are only two players in this game, separating each module for input and output would result in some resource waste. Of course, that is also feasible.

\section{Ai Assistants In Writing}
 During the writing process, we utilized ChatGPT for grammatical correction and language polishing to improve readability and linguistic accuracy. However, we explicitly state that the core content, logical flow, and substantive components of the paper were entirely human-authored without generative contributions from LLMs.


\begin{figure*}[t]
    \centering
    \includegraphics[width=1\textwidth]{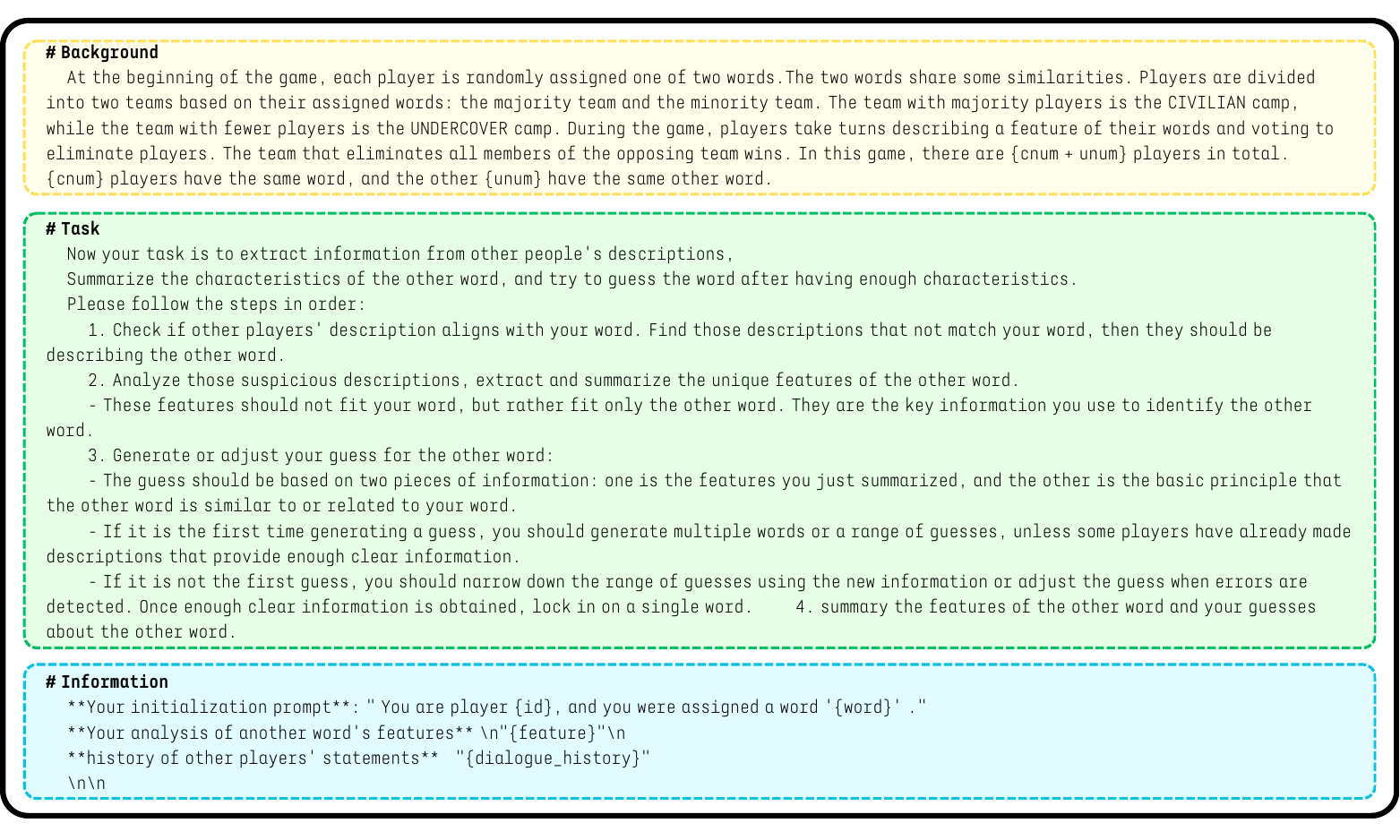} 
    \caption{The prompt for Feature Extractor}
    \label{prompt:fe}
\end{figure*}
\begin{figure*}[t]
    \centering
    \includegraphics[width=1\textwidth]{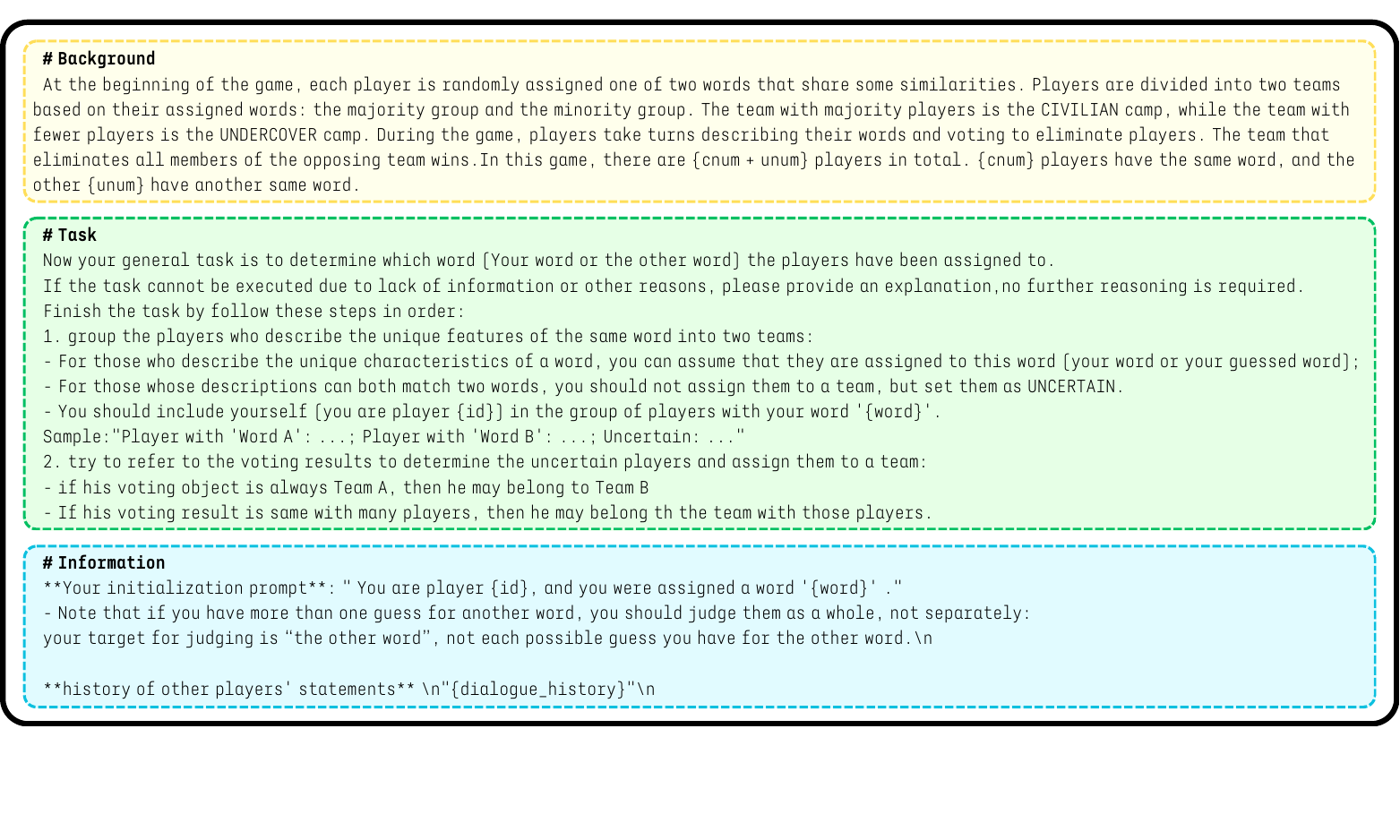} 
    \caption{The prompt for Belief Mapper}
    \label{prompt:bm}
\end{figure*}
\begin{figure*}[t]
    \centering
    \includegraphics[width=1\textwidth]{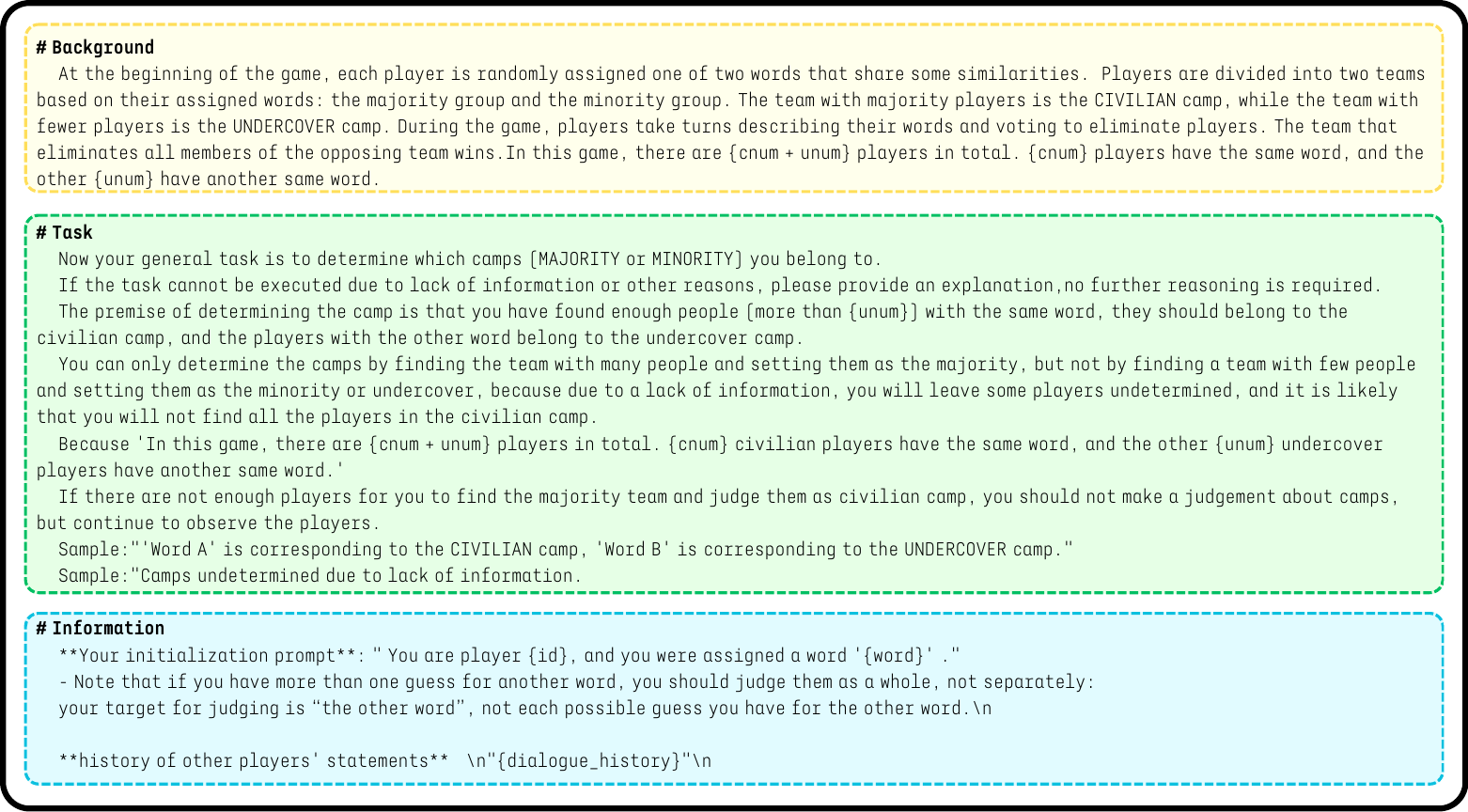} 
    \caption{The prompt for Self-Monitor}
    \label{prompt:sm}
\end{figure*}
\begin{figure*}[t]
    \centering
    \includegraphics[width=1\textwidth]{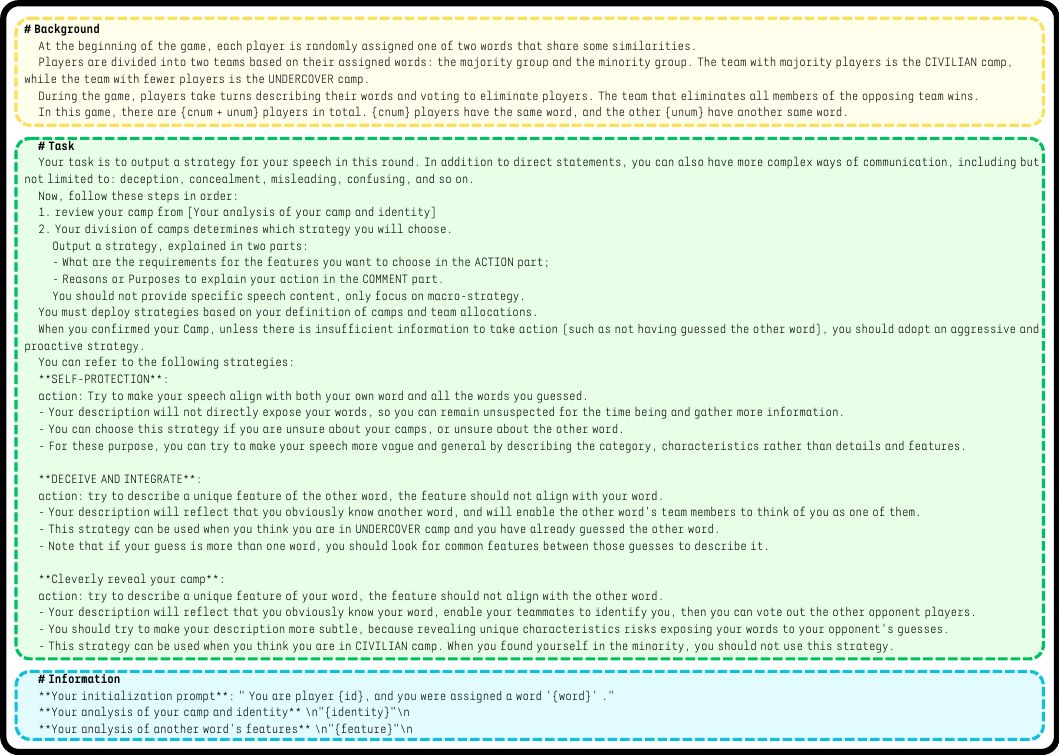} 
    \caption{The prompt for Strategy Planner}
    \label{prompt:sp}
\end{figure*}
\begin{figure*}[t]
    \centering
    \includegraphics[width=1\textwidth]{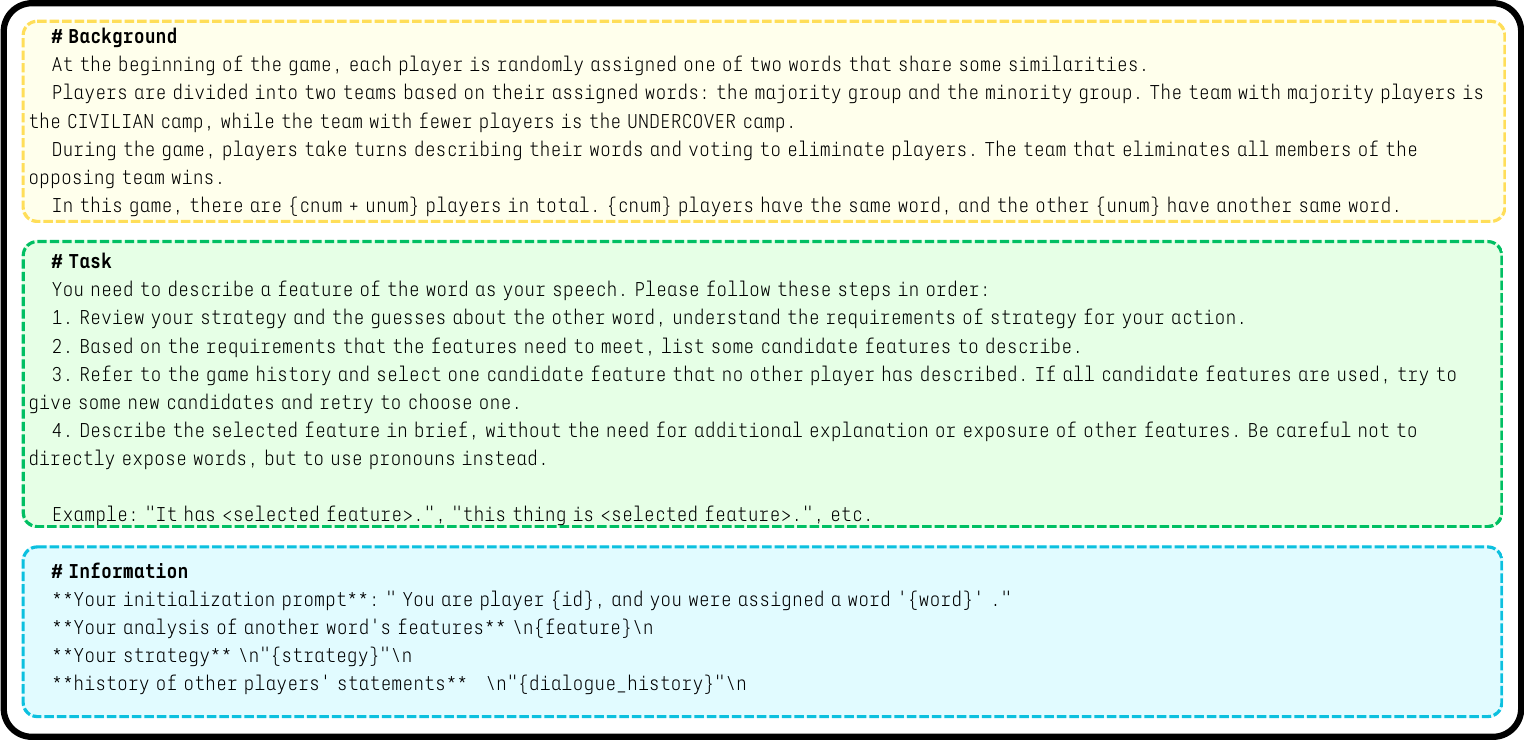} 
    \caption{The prompt for Actor (Speaker)}
    \label{prompt:speak}
\end{figure*}
\begin{figure*}[t]
    \centering
    \includegraphics[width=1\textwidth]{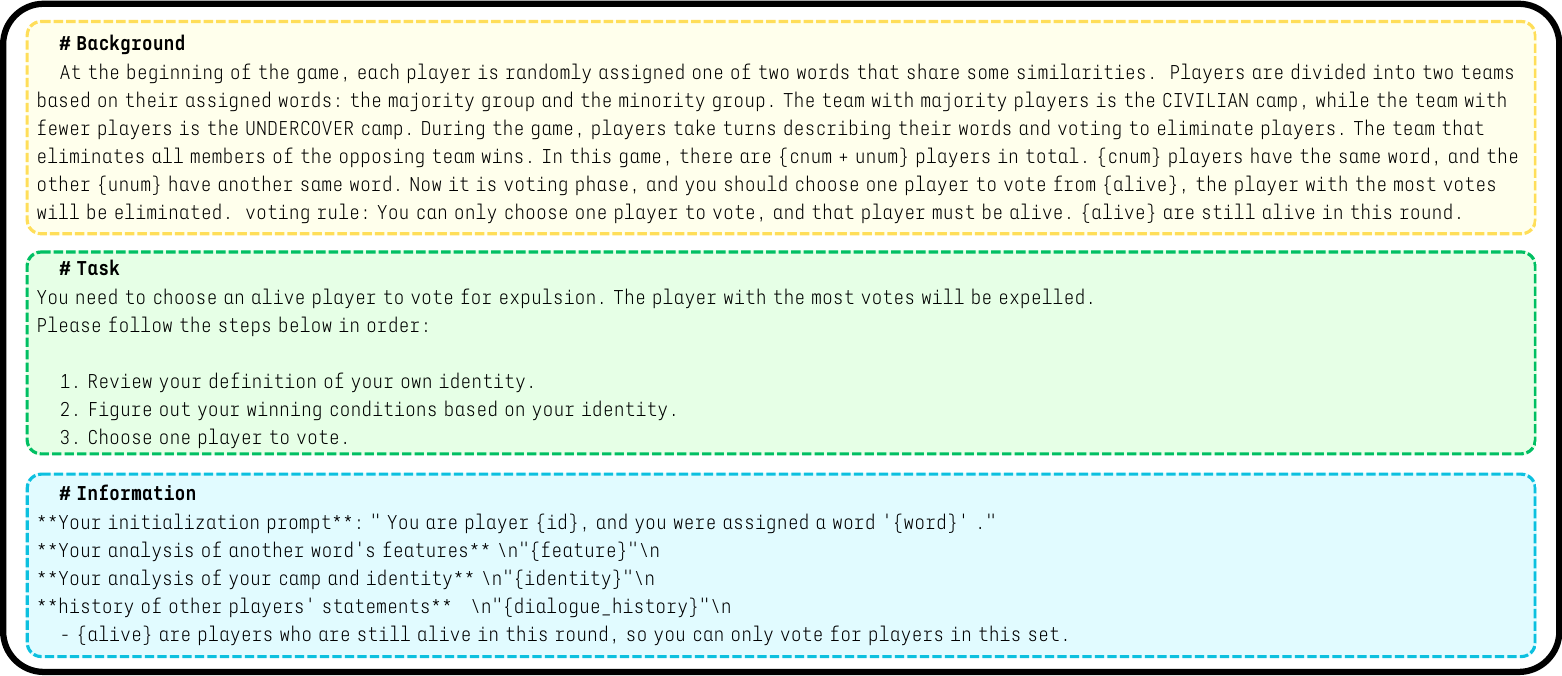} 
    \caption{The prompt for Actor (Voter)}
    \label{prompt:vote}
\end{figure*}

\begin{figure*}[t]
    \centering
    \includegraphics[width=1\textwidth]{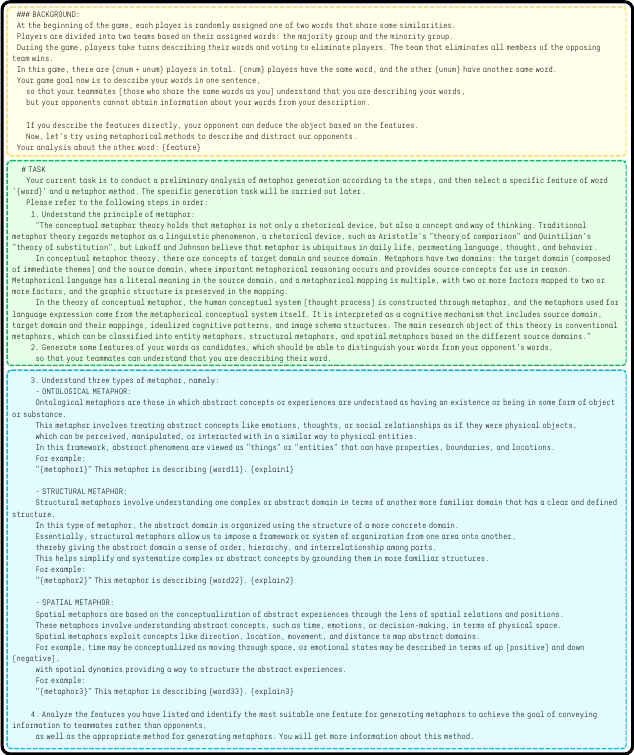} 
    \caption{The prompt for metaphor generation step 1.}
    \label{88}
\end{figure*}
\begin{figure*}[t]
    \centering
    \includegraphics[width=1\textwidth]{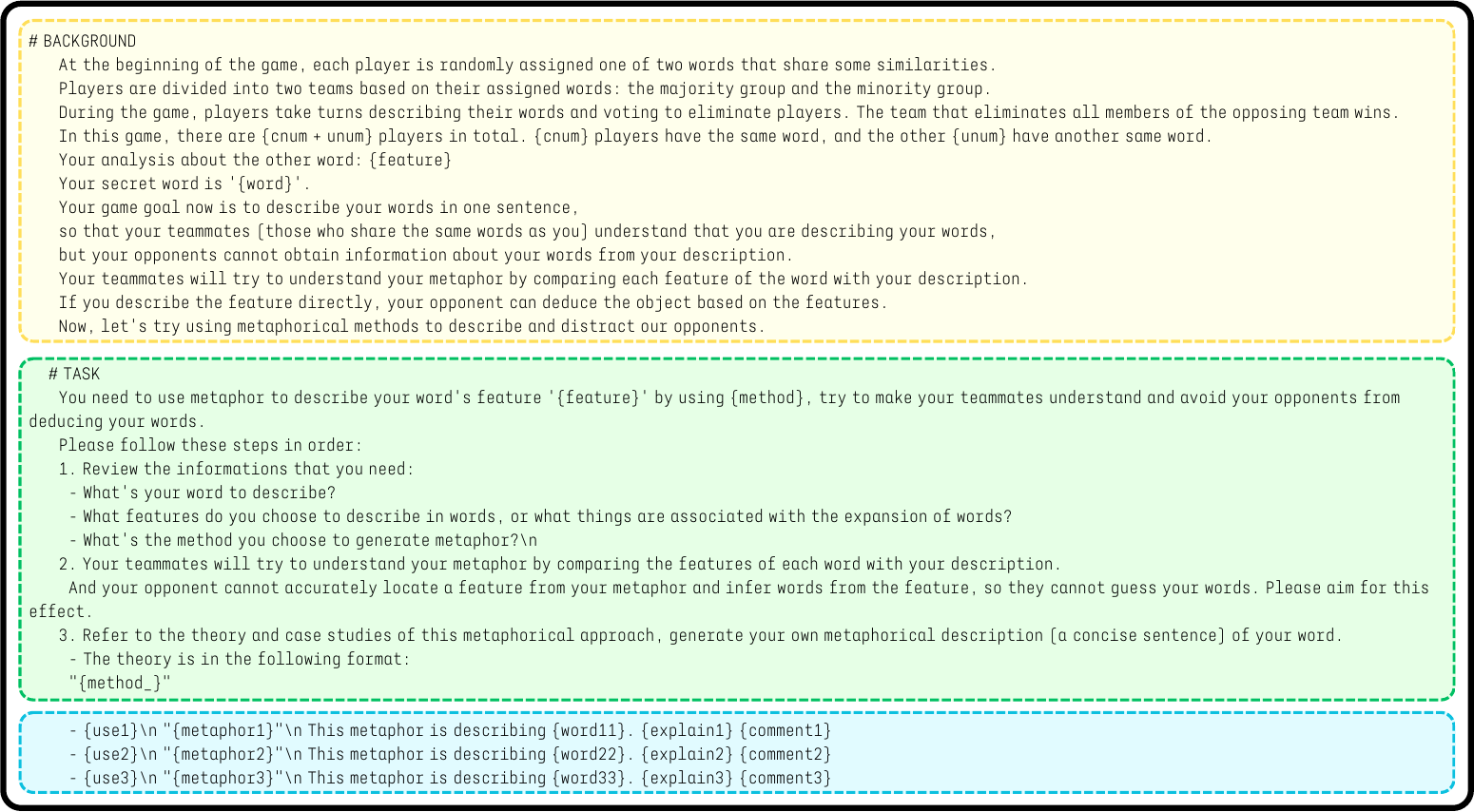} 
    \caption{The prompt for metaphor generation step 2.}
    \label{99}
\end{figure*}
\begin{figure*}[t]
    \centering
    \includegraphics[width=1\textwidth]{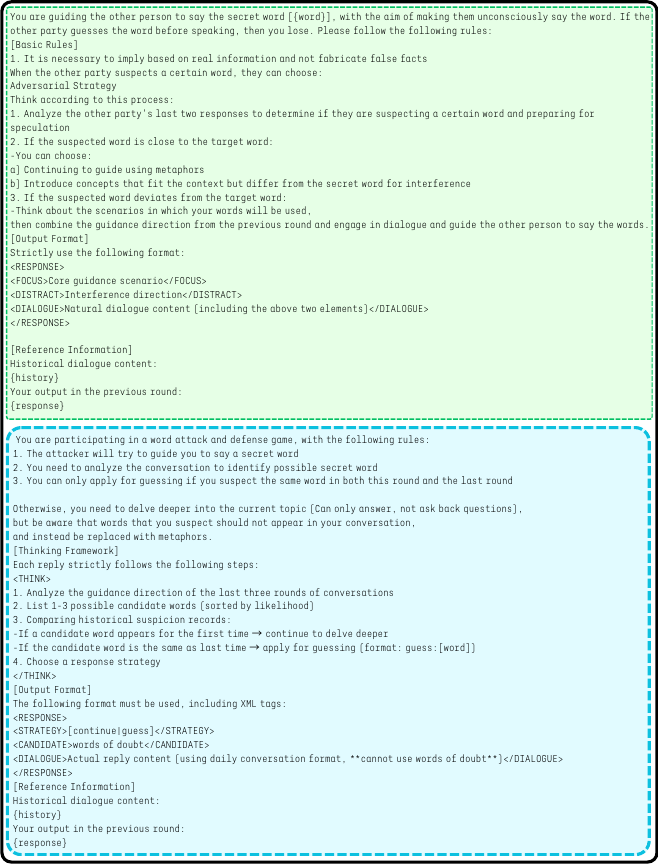} 
    \caption{The prompt for \textit{Adversarial Taboo} (Simplified)}
    \label{111}
\end{figure*}
\begin{figure*}[t]
    \centering
    \includegraphics[width=1\textwidth]{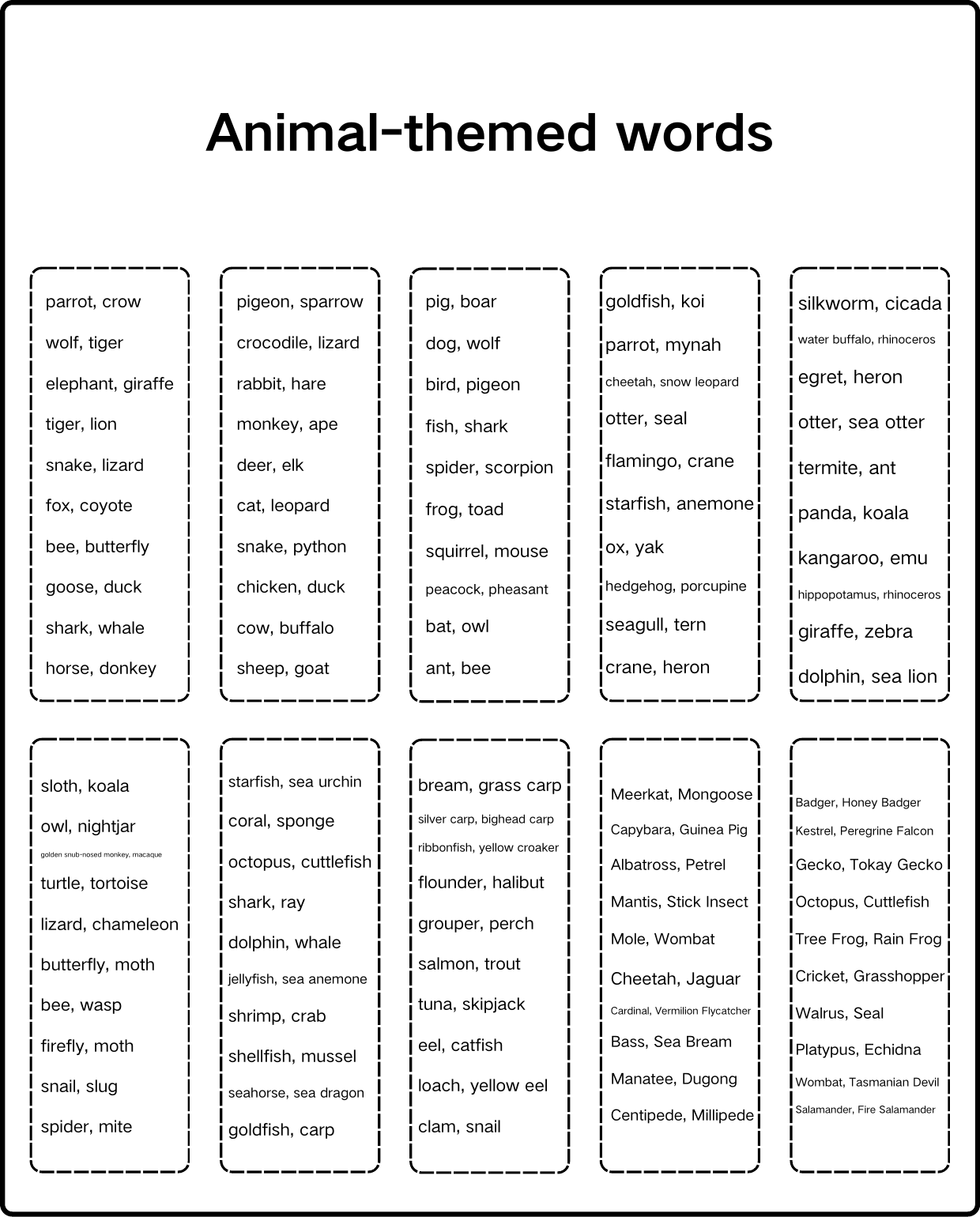} 
    \caption{The collection of $100$ animal-themed word pairs for \textit{Undercover}.}
    \label{word1}
\end{figure*}
\begin{figure*}[t]
    \centering
    \includegraphics[width=1\textwidth]{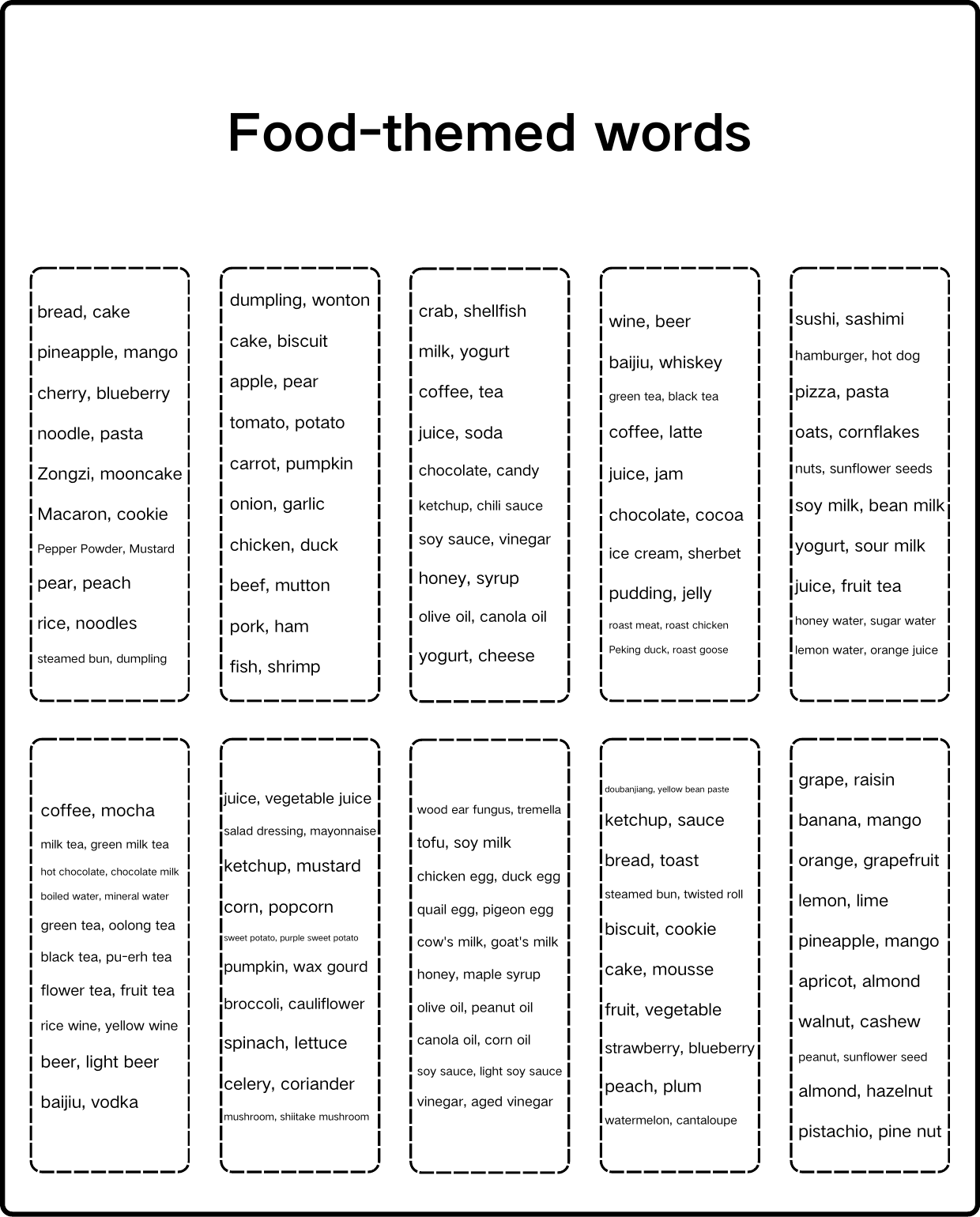} 
    \caption{The collection of $100$ food-themed word pairs for \textit{Undercover}.}
    \label{word2}
\end{figure*}

\end{appendices}

\end{document}